\documentclass{article}

\usepackage[final]{neurips_2022}

\usepackage[utf8]{inputenc} %
\usepackage[T1]{fontenc}    %
\usepackage{url}            %
\usepackage{booktabs}       %
\usepackage{amsfonts}       %
\usepackage{nicefrac}       %
\usepackage{microtype}      %
\usepackage[usenames,dvipsnames]{xcolor}
\usepackage[bookmarks=false]{hyperref}
\hypersetup{
  pdffitwindow=true,
  pdfstartview={FitH},
  pdfnewwindow=true,
  colorlinks,
  linktocpage=true,
  linkcolor=Green,
  urlcolor=Green,
  citecolor=Green
}

\usepackage{natbib}
\usepackage{amsmath}
\usepackage{graphicx}
\usepackage{wrapfig}
\usepackage{subcaption}
\usepackage{enumitem}
\usepackage[normalem]{ulem}
\usepackage{cleveref}

\title{Learning to Navigate Wikipedia\\ by Taking Random Walks}

\author{%
  Manzil Zaheer, Kenneth Marino, Will Grathwohl, John Schultz, Wendy Shang, \\
  \textbf{Sheila Babayan, Arun Ahuja, Ishita Dasgupta, Christine Kaeser-Chen, Rob Fergus} \\
  DeepMind New York\\
  \texttt{\{manzilzaheer, kmarino, wgrathwohl, jhtschultz, wendyshang,} \\
  \texttt{
  sbabayan, arahuja, idg, christinech, robfergus\}@google.com} \\
}

\begin{document}

\maketitle

\begin{abstract}
A fundamental ability of an intelligent web-based agent is seeking out and acquiring new information. Internet search engines reliably find the correct vicinity but the top results may be a few links away from the desired target. A complementary approach is navigation via hyperlinks, employing a policy that comprehends local content and selects a link that moves it closer to the target. In this paper, we show that behavioral cloning of randomly sampled trajectories is sufficient to learn an effective link selection policy. We demonstrate the approach on a graph version of Wikipedia with $38$M nodes and $387$M edges. The model is able to efficiently navigate between nodes $5$ and $20$ steps apart $96\%$ and $92\%$ of the time, respectively. We then use the resulting embeddings and policy in downstream fact verification and question answering tasks where, in combination with basic TF-IDF search and ranking methods, they are competitive results to the state-of-the-art methods.

\end{abstract}

\section{Introduction}
The ability to gather new knowledge about the world is a fundamental aspect of intelligence. Based only on a few words, a fact, question, or even vague idea, humans have the ability to use the internet to find extremely specific information about the world. From the important (how to administer CPR) to the trivial (the LA Raiders won the 1983 SuperBowl), virtually any knowledge is a click away.

In this work, we focus in one particular aspect of this ability: web navigation. In general, navigation is a key component of an embodied agent: the ability to move efficiently toward a target. In known environments, where map information is available, shortest-path algorithms provide a viable solution. However, in novel settings, the agent lacks such global information and must instead \emph{navigate} to the target, using its understanding of the local environment to select actions. 

Other approaches to web agents have focused mostly on retrieval or search engines. These, however,
only provide a partial solution, typically getting close to a desired target but often not exactly to the right page. 
In this work, we present an approach for navigation on graph-structured web data that uses hyperlinks within articles to navigate toward a target. It complements search engines, using them to provide a sensible starting point for a local search for specific information.

\begin{figure}[t!]
    \centering
    \includegraphics[width=0.85\linewidth]{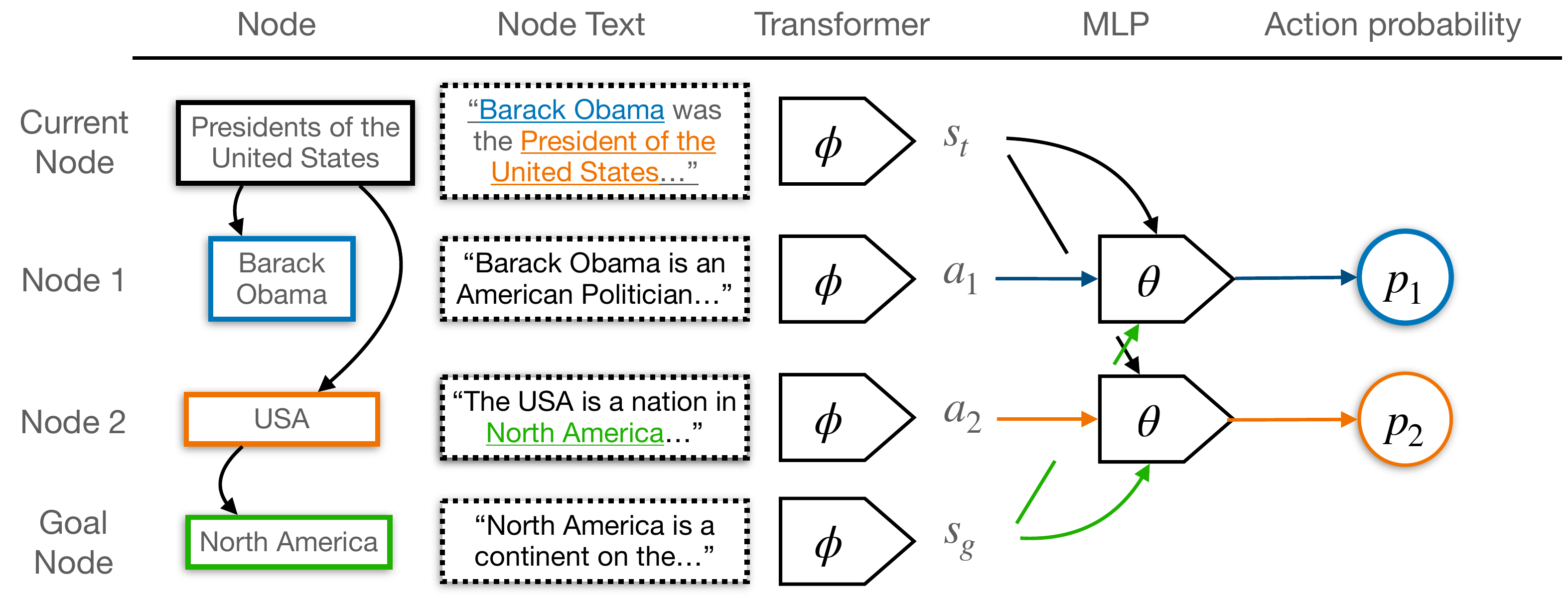}
    \caption{Our agent navigates through hyperlinks on a Wikipedia graph towards a target node. It first embeds the text at each link node using Transformer $\phi(.)$ and then evaluates different actions/links $a_1,a_2$ with a policy $\theta$, conditioned on the current and target embeddings $s_t$ and $s_g$, respectively.}
    \vspace{-3mm}
    \label{fig:teaser}
\end{figure}

To be able to navigate on the web effectively, we must read the text of the page, see how concepts in the current page are related to a possible next page, and use our understanding to learn a policy to make the right navigation decision. 
Consider the scenario shown in Figure \ref{fig:teaser}. The agent is trying to navigate to the target node {\sc North America} while currently at the node {\sc Presidents of the United States} containing two hyperlinks: {\sc Barack Obama} and {\sc USA}. Our approach learns an embedding from the text of the current page and the text for each hyperlink node and passes these to a goal-conditioned policy that selects the best possible action. To make the correct choice in this example, the agent should associate {\sc USA} with {\sc North America}.

The particular web environment we consider here is Wikipedia, converted into a graph form. %
Each of the 38M paragraphs is represented by a node; edges are links within and between articles. We introduce an approach for navigation on this graph based on behavioral cloning of automatically generated trajectories. This allows for unsupervised pre-training of both web page embeddings and navigation policy. We explore different trajectory distributions for training the model, the best practical choice being equivalent to random walks on the graph.

{\noindent \bf Contributions:} Our work presents the first viable solution to the problem of web navigation\footnote{A demo of our agent navigating is shown here \url{https://www.youtube.com/watch?v=LVqOaPKpC2c}}, as demonstrated by reliable (>90\% success) navigation on a graph of size $10^7$. Previous attempts such as \citet{Nogueira16} show a far inferior performance, well below that needed for practical utility. Furthermore, as there is nothing Wikipedia-specific about our approach and it scales gracefully, the method is applicable to general web settings. We also demonstrate the relevance of the pre-train + fine-tune paradigm to the web domain, showing that pre-training for navigation is a suitable objective to aid downstream tasks with limited data such as Q\&A. 
We demonstrate this by applying our navigation approach on two challenging tasks: fact verification (FEVER;~\cite{fever}) and Q\&A (Natural Questions;~\cite{orqa}) for evidence gathering. 
In both settings, we show a jump in performance over existing search-based approaches that rely on retrieval followed by re-ranking, matching state-of-the-art methods despite using far simpler retrieval and re-ranking mechanisms.

\section{Background}
\label{sec:background}

\textbf{Wikipedia as a navigation environment:}
Human behavior on Wikipedia \citep{West12} was collected via the Wikispeedia game \citep{Wikispeedia}, which asked the players to move between a pair of arbitrarily chosen articles in the fewest steps. This rule allows the emergence of a semantic distance between concepts that facilitates retrieval. \citet{West12a} conducted initial studies on this dataset for navigation, using supervised learning and reinforcement learning methods on a small subset of articles ($4.6$K) with simple TF-IDF features. 

Most relevantly, \citet{Nogueira16} explore the same problem of navigation on a large Wikipedia graph ($12$M nodes). They train LSTM agents to perform navigation on the graph, using Bag-of-Word features for each article. The trained agent is then applied to a Jeopardy Q\&A task. Our approach revisits this framing but with a number of important differences: (i) training on a significantly larger graph ($38$M vs $12$M nodes, $51.5$M vs $380$M edges); (ii) using Transformer-based architectures for paragraph encodings and for the navigation policy; (iii) they use a single fixed start node, whereas ours is able to navigate from any start node; (iv) removal of beam search, or other complex search mechanisms at evaluation time. 
Collectively, our approach boosts the navigation success rate from $12.5\%$ in~\citet{Nogueira16} to over $90\%$,\footnote{Exact comparisons are infeasible given idiosyncrasies in graph construction and other differences in setup.} despite a much larger graph.

\textbf{Fact verification and Q\&A using Wikipedia:}
Wikipedia is a rich resource for evaluating language understanding through Q\&A tasks.  \citet{Chen17} combine TF-IDF retrieval with an LSTM document reader to find answers to free-form questions within Wikipedia documents. Contemporary Q\&A systems \citep{Semnani20, Qu20} extend this general approach to more complex pipelines, combining traditional and neural methods with inverted index search/retrieval schemes, followed by document readers that perform re-ranking. We demonstrate our navigation approach in conjunction with very basic retrieval and re-ranking approaches, so as to clearly demonstrate the value of navigation. In common with our approach, \citet{Wang21} also use a graph version of Wikipedia, but with the goal of aligning it to a knowledge base, rather than navigation. \citet{stammbach-neumann} propose a scheme for finding evidence for the FEVER \citep{fever} benchmark that explores all links one step away from an initial retrieval set. This can be viewed as a 1-step navigation operation, in contrast to our scheme that permits an arbitrary number of steps.

\textbf{Navigation in knowledge bases:}
Graph navigation is a key element in utilizing knowledge bases (KB), where facts are represented as tuples of entity nodes connected by a labelled edge corresponding to relation between the two nodes. 
For example, \citet{Das17} answer complex queries by navigating the graph, inferring missing edges along the way, and finding a valid path to the answer node.
In our case, the graph is fixed but consists of natural language text and is partially observed. This makes the task much harder on two fronts. First, the policy must comprehend the natural language actions, without access to crisply labelled relation actions. Second, the action space is open and varying in size, unlike in a KB which typically has a fixed schema, i.e. relations/actions come from a fixed size set.
Perhaps the closest work involving natural language actions, ~\citet{uscpaper} augment a KB with textual nodes. However, it still operates under the fixed KB schema.
Further improvements on navigation in KB have been shown by incorporating tricks like reward-shaping~\citep{deeppath} and action dropout~\citep{victoria}.
We adopted the action dropout where some outgoing edges are masked during training to enable more effective path exploration.
Pushing more on exploration, inspired by AlphaGo, \citet{jianfenggao,He22} combine the policy with Monte-Carlo tree search to navigate KB. 
We find that no search method is necessary for our approach at evaluation time, with good performance obtained from the policy alone.

\textbf{Internet powered Q\&A:}
Several recent works have attempted complex question answering using the internet. \citet{Talmor18} retrieve snippets of information from the web with a search engine to answer a structured decomposition of the question with a simpler Q\&A model. The recent WebGPT \citep{WebGPT} system takes a broadly similar approach to long-form question answering. WebGPT accures information from the Bing search API as well as some navigation, and passes it to an LM to generate the answer text. The complexity of the system necessitates the use of human demonstrations for training. 
Related work from \citet{Lazaridou22} utilizes a few-shot supervision of an LM to perform Google queries, which are then used by the LM to generate an answer. They show that by incorporating information from Google queries, the answers are more factual than those generated directly from LMs.
Another line of work, has leveraged link structure of webpages found on the internet to improve Q\&A. In LinkBERT~\citep{yasunaga2022linkbert} is pretrained to capture dependencies between documents and results in improvement for for multi-hop reasoning and few-shot reading comprehension.
\citet{Asai2020Learning} use the link structure to enhance open domain Q\&A by learning to retrieve according to reasoning paths.
In contrast to these works, we focus on mastering navigation rather than any combination with search/retrieval, albeit in more limited domain (Wikipedia vs the internet). However, our approach could potentially act as a core component in a general web navigation, being more versatile on account of not requiring supervision.

\section{Approach}\label{sec:approach}
We consider navigation in a web environment $\mathcal{W}=\{\mathbf{N},\mathbf{L}\}$, where $ \mathbf{N} := \{n_i\}_{i=1}^N$ is the set of nodes, each $n_i$ representing a web page with content $s_i$. The links between pages are represented by $\mathbf{L}:= \{l_i\}_{i=1}^N$, where $l_i=\{l_i^1,\ldots,l_i^j,\ldots,l^{M_i}_i\}$ are the set of $M_i$ (directed) out-links from node $i$. We define $\text{NH}(n_i)$ as the set of nodes reached by following all links $l_i$ from node $n_i$, where the $j$th neighbor is given by $l_i^j(n_i)$.

We assume some target node $n_g$ and some starting node $n_0$ are given, and wish to train an agent that can navigate from $n_0$ to $n_g$ by walking along the edges of the graph within $T$ steps. To accomplish this, we construct a parametric policy $p_\theta(n_{t+1} | n_{t}, n_g)$ which, at time step $t$, parameterizes a distribution over $n_{t+1} \in \text{NH}(n_t)$. We train this model via behavioral cloning to maximize%

\begin{align}
    \mathcal{L}(\theta) = \mathbf{E}_{\mathcal{E}(n_0, \ldots, n_{T})}\left[ \sum_{t=0}^{T-1}\log p_\theta(n_{t+1} | n_{t}, n_g)  \right]
    \label{eq:bc}
\end{align}
where $\mathcal{E}(n_0, \ldots, n_{T})$ is our trajectory distribution which generates trajectories $(n_0,\ldots, n_T)$ that we would like our model to replicate (i.e. setting $n_g=n_T$).  

\subsection{Trajectory distribution} 
\label{sec:trajdist}
This framework opens up many choices for $\mathcal{E}$. Clearly some will result in ineffective navigation strategies while others may enable us to train a policy which is capable of generalizing to new graphs and downstream applications. We elaborate on some choices below.

\textbf{Reverse trajectories:}
We define $\mathcal{E}(n_T) = \text{Uniform}(\textbf{N})$ (the uniform distribution over nodes in the graph). Now we can define
\begin{align}
    \mathcal{E}(n_t|n_{t+1}) = 
    \frac{1}{|\{n_i;  n_i \in \text{NH}(n_t)\}|} \quad  \text{ if } n_{t+1} \in \text{NH}(n_{t}), \quad \text{ else } 0  
\end{align}
or, a distribution which uniformly walks the \emph{reverse} graph (with transposed adjacency matrix). We begin with some uniformly chosen target node and walk backward randomly to $n_0$ in such a way that we can reach the target by walking forward from $n_0$.

Intuitively, training with the reverse trajectories defined above should allow our model to be able to reach any node in the graph from any node it is connected to. Unfortunately, in realistic web graph data, there exist many ``dead-ends'' or nodes with no in-going edges. The trajectories get stuck at these nodes leading to low-diversity training data. 

\textbf{Random forward trajectories:}
In our real-world knowledge graph data, there are far fewer nodes which have no out-going edges. Thus we propose an alternative method to generate target nodes based on randomly walking forward. We define $ \mathcal{E}(n_0)$ to be the $\text{Uniform}(\mathbf{N})$ and 
\begin{align}
     \mathcal{E}(n_{t+1}|n_{t}) = 
    \frac{1}{|\text{NH}(n_{t})|} \quad \text{ if } n_{t+1} \in \text{NH}(n_{t}), \quad \text{ else } 0.
\end{align}
or, a distribution which uniformly picks a starting node and randomly walks the \emph{forward} graph for $T$ steps. Training in this way should encourage our model to be able to reach any node which is reachable from some uniformly chosen start. We note that the target node distribution $\mathcal{E}(n_T)$ will not be uniform and instead be a function of the graph structure. We also note that while this generates a $T$-step trajectory, because it is generated by random walk, the shortest path distance between start and target nodes may be smaller.

\textbf{Random shortest paths:}
We define $ \mathcal{E}(n_T) = \text{Uniform}(\textbf{N})$ and let $ \mathcal{E}(n_0|n_T)$ be the uniform distribution of nodes whose shortest path length to $n_T$ is $T$ steps. We then define $ \mathcal{E}(n_t|n_0, n_T)$ to be the delta distribution that $n_t$ is the $t$-th node of this shortest path. While it can be challenging to compute the probabilities of this distribution efficiently we can draw samples from it using a shortest path finding algorithm such as Dijkstra's algorithm.

\subsection{Model parameterization}
From above, we can see the core of our model is the $p_\theta(n_{t+1}|n_t, n_T)$ distribution. Given a sampled trajectory that begins at $n_0$, this is the distribution which attempts to choose the correct action to take at timestep $t+1$ starting at node $n_t$ in order to get to target node $n_T$. We parameterize this distribution as a function of the text available at node $n_t$. %
\begin{align}
    p_\theta(n_{t+1}|n_t, n_g) \propto \exp(f_\theta(s_{t+1}, s_t, s_{g}, t))
\end{align}
where $f_\theta$ embeds the text $s_g$, $s_t$ and $s_{t+1}$ of the target, current and destination nodes, respectively. Further details on how we apply this general model to Wikipedia are given in Section~\ref{sec:model} and an alternative interpretation of this approach based on variational inference in a latent-variable model is presented in Appendix \ref{app:varinterp}.

\section{Application to Wikipedia}

\subsection{Navigation and sentence search tasks}
\label{sec:tasks}

\textbf{Data processing:} We convert a snapshot of English Wikipedia into a graph but split each article ($\bar{\mu}=1000$ words) into paragraph-sized blocks ($\bar{\mu}=100$ words), which form the \emph{nodes}. \emph{Edges} in the graph are the union of (i) organic hyperlinks, (ii) additional entity linking and (iii) next/previous paragraph links. 
We consider two snapshots from year 2017 and 2018, which produces large full graphs with $\sim 37$M nodes and $\sim 370$M edges. 
For initial experiments, we also use a smaller subsampled graph, for which we sub-sample disjoint sets of 200k train / 200k evaluation nodes from the 2018 graph. 
All the obtained graphs are well connected with median path length of 15.
Further details have been relegated to the Appendix 
(full graphs in Appendix~\ref{sec:appendix_data}, the 200k graph in Appendix~\ref{sec:small_graph}, and graph statistics in Appendix~\ref{sec:wiki_stats}).

\textbf{T-step navigation:}
The most basic task is $T$-step navigation. At the start of the episode, we randomly sample a start node and generate a $T$-step random walk trajectory. We then take the last node of this trajectory as the target $n_g$. The agent is then given a time budget $B$ (in most experiments, we fix max steps to $B=100$), and 
succeeds
if it reaches the target node within $B$ steps. In our experiments, we show results on various graphs and splits with $T=\{5,10,20\}$-step navigation.

\textbf{1-T Multistep navigation:}
The Multistep task is much like the previous setting, except that instead of generating fixed $T$-step trajectories, we sample from $\text{Uniform}(1,T)$, $T=20$. 

\textbf{T-step sentence search:}
This is the same setup as $T$-step navigation, except that rather than provide the full text for the target node $n_g$, only one sentence within that text is selected at random to compute the target embeddings from. Thus agents have the more challenging task of locating the correct node, given only a small snippet of its text (see Section~\ref{sec:model} for details on target representation). We include this task because one of the difficulties of web navigation is establishing which page should be the target in the first place. It also has direct relevance to our downstream fact verification task.

\subsection{Model architecture}
\label{sec:model}
\textbf{State representation:} For each node $n_i$ we embed its text $s_i$ using a Transformer $\phi(.)$ to produce $\phi(s_i)$. By basing the state representation on semantic content, rather than i.e. node index, the agent is able to learn about the relationships between entities in a manner that allows for generalization to new graphs at evaluation time, instead of memorizing the structure of the training graph. 
Following~\citet{karpukhin20dpr}, we append to the node text the title of the Wikipedia article to which it belongs. This often provides context to the node that might otherwise be missing -- in many articles, the subject of the article will be referred to indirectly (e.g. it might state that ``she'' was born in 1945 instead of repeating the subject's name).

For those experiments where the node embeddings are kept fixed, $\phi(.)$ is a pre-trained RoBERTA \citep{Roberta} model that encodes the node text \& title. However, in our large-scale experiments, we pre-train the Transformer $\phi(.)$ directly using the text \& title (see Appendix \ref{sec:will_pretrain}).
In both cases, we apply the Transformer to the tokenized text, take the mean over the input tokens, and apply a $\texttt{tanh}$ nonlinearity to produce $\phi(s_i)$.

\textbf{Target representation:} 
Similarly, we use the article text $s_g$ of the target node as input to the Transformer model $\phi$ to compute its representation $\phi(s_g)$. For some tasks, such as sentence search or fact verification, the target representation instead relies on a single sentence from each node or other text specific to the task and a separate transformer model $\phi_\text{target}$ is trained to embed the target.

\textbf{Navigation policy network:} The network operates on the target node $n_g$ and the current node $n_t$, and outputs a distribution over the possible actions that can be taken at $n_t$. As the nodes in the graph have a variable number of outgoing edges, the policy must produce a distribution with a variable number of outcomes. To do this, we first concatenate the embeddings of $s_g$ and $s_t$, and pass them through a 1-layer feed-forward network to produce a combined embedding $e_{tg} = FF[\phi(s_t),\phi(s_g)]$. Next we embed each possible action $a_i$. 
When we use a fixed $\phi$,
we directly use $a_i = \phi(s_i)$ for $n_i \in \text{NH}(n_t)$. 
However this is not feasible when pre-training our embedding model due to memory concerns. Instead, we compute $a_i=\phi(s_t[l^i_t])$
where $s[l^i]$ selects the words in $s$ which belong to hyperlink $i$ -- see Appendix \ref{sec:will_pretrain} for more details. We then define the probability of moving to $n_i$ from $n_t$ with target $n_g$ as $p(n_i | n_t, n_g) \propto \exp{(e_{tg}\cdot a_i)}$ which is normalized to produce a distribution over the $|\text{NH}(n_t)|$ actions.

In our larger-scale model, we also include the trajectory leading up to the current node $(n_0,\ldots, n_t)$. We pass their embeddings, combined with the target embedding, $[\phi(s_0), \ldots, \phi(s_t), \phi(s_g)]$, through a 4-layer transformer and use the output as $e_{tg}$ when computing the transition probabilities above. We explore these two policy architecture choices, feed-forward and Transformer in Table \ref{tab:fullnavigation}.

\subsection{Downstream tasks}
Finally, we use our navigation approach to gather information for the tasks of fact verification on the FEVER benchmark \citep{fever} and question answering on Natural Questions (NQ)~\citep{orqa} which also use Wikipedia as a knowledge base. 
One important difference from our earlier navigation task is that the target node $n_g$ is not specified at evaluation time. Instead, we are given a claim whose veracity must be established or a question that must be answered by finding information in a particular node in the graph. To determine which node this is, we use a target encoder which maps the claim or question to an embedding vector that can be used in place of $\phi(s_g)$ in Section \ref{sec:model}. %

Due to the limited size of the downstream training sets, we pre-train our graph embeddings, navigation policy and target encoder on the similar sentence search task. We then freeze the embeddings and policy and fine-tune the target encoder on $(n_g,s_g)$ pairs from the downstream task's training set. 
To avoid over-fitting, we add an auxiliary loss which enforces the similarity between our target embedding and the embedding of the ground truth node containing the sentence needed to resolve the claim or question. Training and evaluation on FEVER and NQ required alignment between our version of Wikipedia and the one used to compile the benchmark and details of this can be found in Appendix \ref{sec:fever_eval} and \ref{sec:nq_eval}.

Next, our navigation method requires a starting node. To put our agent close to the articles we need in the downstream tasks, we run a popular variant of TF-IDF, called BM25~\citep{bm25}. We first create an index over all of the nodes in the graph, then take the top-5 BM25 matches as the starting nodes, and run navigation for $20$ steps on each. Note that BM25 often does not find the exact text (i.e. the hits@1 is not  high) but matches are in the right vicinity, usually the same or similar article. 
Thus the agent only has to navigate a few steps to reach the target: in FEVER and NQ datasets, the mean length of shortest path between start and target node is $4.3$ and $5.1$ steps respectively. 

The final component is a re-ranker that takes all the sentences from all nodes visited by the agent and assigns a match score between each one and the claim. The ranked list of sentences can then be used for benchmark evaluation. 
We explore two types of ranker: (a) basic TF-IDF and (b) a transformer based model like BERT or BigBird, fine tuned using hard negatives mined from the TF-IDF ranker.
See Appendix \ref{sec:fever_eval} and \ref{sec:nq_eval} for more details.

Evaluation of our approach thus consists of: (i) given a claim/question, use BM-25 to return a list of promising start locations; (ii) from the top $N=10$ of these, run our navigation model for $20$ steps, using the target encoder to embed the claim/question, (iii) use the re-ranker to score sentences visited against the claim/question and (iv) select the top 5 for computing the recall, precision, or F1 score, for comparison on the benchmarks.

\section{Experiments}
\subsection{Investigation of training trajectory distribution }
\label{sec:investtrajectory}

On the smaller 200k node graph, we investigate different choices for the trajectory policy distributions $\mathcal{E}$ described in Section~\ref{sec:approach}: (i) \textbf{Random Forward Trajectories}, where we randomly sample start nodes and do a random walk to get a target, (ii) \textbf{Reverse Trajectories}, where we randomly sample a target node and and do a random walk on the reverse graph to get a start node, and (iii) \textbf{Random Shortest Paths} where we randomly sample a start node and take a random walk and then compute the shortest path between the start and target nodes. For each of these we trained policies on $5$, $10$, and $20$ step navigation tasks. We evaluate these in two different ways: (i) forward sampling of source and target in the disjoint evaluation graph, (ii) reverse sampling of source and target. The results of this experiment are shown in Table~\ref{tab:expertpolicy}. Unsurprisingly, we can see that forward and reverse trained policies both do better when evaluated in the way that matches training. However, forward trajectory policies generalize much better to reverse navigation. When we analyze the traces of these experiments, we see that in the reverse sampling, many target nodes are in ``dead-ends'' where many target nodes do not have long random trajectories to sample. This results in easier navigation (as we can see from higher accuracy in reverse trajectory trained policies on reverse navigation) and more extreme overfitting, leading to poor performance of reverse trajectory on forward navigation.

Not surprisingly, we also see that training on a shortest path trajectories leads to better performance both in the forward and reverse sampling. This advantage is especially seen in the $20$-step navigation task. Ideally, we would always train on shortest paths. However, the time complexity of pre-computing all shortest paths in a graph using Dijkstra's algorithm is $O(V(E+VlogV))$ for the number of nodes $V$ and number of edges $E$ in the graph. 
On the full $38$M Wikipedia graph, this computation would be completely intractable. Luckily, the random forward trajectories perform almost as well and can be computed in constant time with respect to the size of the graph and only linear time with respect to the trajectory length. For all subsequent experiments, we use the \textbf{Random Forward Trajectories} for our trajectory policy and evaluate on the forward version of the navigation tasks, where we call our approach \textbf{Random Forward Behavioral Cloning (RFBC)}.

\begin{table}[t]
  \vspace{-5mm}
  \caption{Different trajectory distributions on small graph (200k nodes)}
  \label{tab:expertpolicy}
  \label{pretraining}
  \centering
  \begin{tabular}{@{}lcccccc@{}}
    \toprule
    \multicolumn{1}{c}{} &
    \multicolumn{3}{c}{Navigation (Forward)} & 
    \multicolumn{3}{c}{Navigation (Reverse)} 
    \\
    \cmidrule(r){2-4}
    \cmidrule(r){5-7}
    Trajectory Policy    & 5 & 10 & 20 & 5 & 10 & 20 \\
    \midrule
    Forward & 85.3  & 76.4 & 67.5 & 31.1 & 32.6 & 19.9 \\
    Reverse & \phantom{0}2.3 & \phantom{0}0.5 & \phantom{0}0.2 & 91.3 & 87.9 & 87.3  \\
    Shortest path & 86.7 & 85.0 & 84.6 & 74.3 & 31.6 & 28.4 \\
    \bottomrule
  \end{tabular}
  \vspace{-3mm}
\end{table}

\subsection{Navigation}

We next do a more thorough set of experiments on the smaller $200$k Wikipedia graph. Because the graph is much smaller than the full Wikipedia and thus prone to overfitting, for all of our methods and baselines we use the fixed RoBERTA embeddings $\phi(s_i)$, the simplest 1-layer feed-forward network, and MiniBERT~\citep{minibert} for target sentence embedding $\phi(s_g)$.

Alongside our approach, we employ the following baselines/ablations:
\begin{itemize}
\setlength{\itemindent}{-2em}
\setlength{\itemsep}{-0.1em}
\item RFBC: Random-forward behavioral cloning (our approach). 
\item RFBC + RF: RFBC but with random features sampled from the unit sphere instead of $\phi$
\item RL + RF: RL with random features.
\item Random: policy that chooses random action (out-link) at each timestep.
\item Greedy: selects the action embedding which has the smallest cosine distance with the target.
\item Random DFS: DFS of depth equal to number of steps, choosing actions at random. 
\item Greedy DFS: DFS but select action with smallest cosine distance to the target. 
\end{itemize}
More training details are in Appendix~\ref{app:smallgraph} and computational cost of each in Appendix~\ref{sec:efficiency}.

The results evaluated on the held-out graph are shown in columns 1-4 of Table~\ref{tab:simplenavigation}. Our approach (RFBC) performs well, achieving 77\% in the multistep case. Performance also drops as the distance to the target increases. In contrast, the RL agent performs poorly (around $40$\% on navigation tasks). Because we generate virtually infinite expert trajectories with RFBC, RL performs worse as it is trained with essentially the same amount of exploration, but with a more sparse signal than behavioral cloning. Using random features in place of $\phi(.)$ caused the performance to drop to near chance, showing that the models are utilizing the semantics at each node and are not relying on a general search strategy of some kind. Random methods also do quite poorly, even when adding a DFS. Greedy methods do a bit better than random, especially with DFS, but still well below our method. 

We also consider the more challenging sentence search task described in Section~\ref{sec:tasks}. This task requires not only finding a known target node, but learning to find a target node given only a single random sentence from that node, more closely matching our downstream task and the way humans tend to approach finding information on the web. Columns 5-7 of Table~\ref{tab:simplenavigation} show performance on this task for our approach and RL. Our method has reduced performance on the task due to the extra difficulty of learning the target embedding, but still performs reasonably. RL, however, degrades to near chance performance.

\begin{table}
  \vspace{-5mm}
  \caption{Small graph navigation (200k nodes) -- Success rate (\%)}
  \label{tab:simplenavigation}
  \centering
  \begin{tabular}{@{}lccccccc@{}}
    \toprule
    \multicolumn{1}{c}{} &
    \multicolumn{4}{c}{Navigation} & 
    \multicolumn{3}{c}{Sentence Search}
    \\
    \cmidrule(r){2-5}
    \cmidrule(r){6-8}
    Method     & 5 & 10 & 20 & multistep & 5 & 10 & 20 \\
    \midrule
    RFBC (ours) & 85.3  & 76.4 & 67.5 & 77.4 & 60.7 & 47.6 & 34.6 \\
    RL & 41.4  & 40.2 & 41.6 & 43.6 & 14.0 & \phantom{0}1.5 & \phantom{0}8.9 \\
    RFBC + RL (ours) & 85.1 & 77.6 & 68.1 & 78.3 & - & - & - \\
    RFBC + RF & 17.4 & 16.3 & 16.6 & 17.1 & - & - & - \\
    RL + RF & \phantom{0}6.0 & \phantom{0}7.5 & \phantom{0}7.0 & \phantom{0}6.7 & - & - & - \\
    Random & 12.3  & 14.9 & 12.3 & 14.0 & - & - & - \\
    Greedy & 19.7& 16.7 & 21.7 & 23.4 & - & - & -  \\
    Random DFS & 10.0 & \phantom{0}9.5 & \phantom{0}8.3 & 10.0 & - & - & -  \\
    Greedy DFS & 31.1 & 23.8 & 22.7 & 51.8 & - & - & -  \\
    \bottomrule
  \end{tabular}
  \vspace{-4mm}
\end{table}

\subsection{Navigation on full 38M node Wikipedia graph:}
\label{sec:fullwikiexp}
\vspace{-2mm}
Our method scales naturally to much larger graphs, which we demonstrate by training on the entire Wikipedia 2017 graph. We evaluate on the entire Wikipedia 2018 graph. 
Each graph is $\sim37$M nodes and  $\sim370$M edges (see Table~\ref{tab:graphsize} for details). 
We would like to point out there is significant evolution (difference) between the 2017 and 2018 graph as analysed in Appendix~\ref{sec:wiki_stats} along with further statistics.

\begin{table}[t]
    \centering
    \caption{Wikipedia graph statistics}
    \label{tab:graphsize}
    \begin{tabular}{@{}cccccc@{}}
    \toprule
    Year & \# Articles & \# Nodes & \# Edges & \# Words / Node & Median Path Length \\
    \midrule
    2017     & 4.92M & 36.3M & 359M & 110 & 15\\
    2018     & 5.27M & 38.5M & 387M & 109 & 15\\
    \bottomrule
    \end{tabular}
    \vspace{-4mm}
\end{table}

We explore both architectures described in Section~\ref{sec:model} for the policy network: (i) the single feed-forward layer (as used in the smaller graph) and (ii) the 4-layer Transformer model. We also compare using a fixed RoBERTA model for $\phi$ versus pre-training a text transformer directly on the navigation task (see Appendix \ref{sec:full_wiki_nav}). DistillBERT~\citep{Sanh2019DistilBERTAD} is used for $\phi(n_g)$. 

\begin{figure}[b!]
    \centering
    \includegraphics[width=1.0\linewidth]{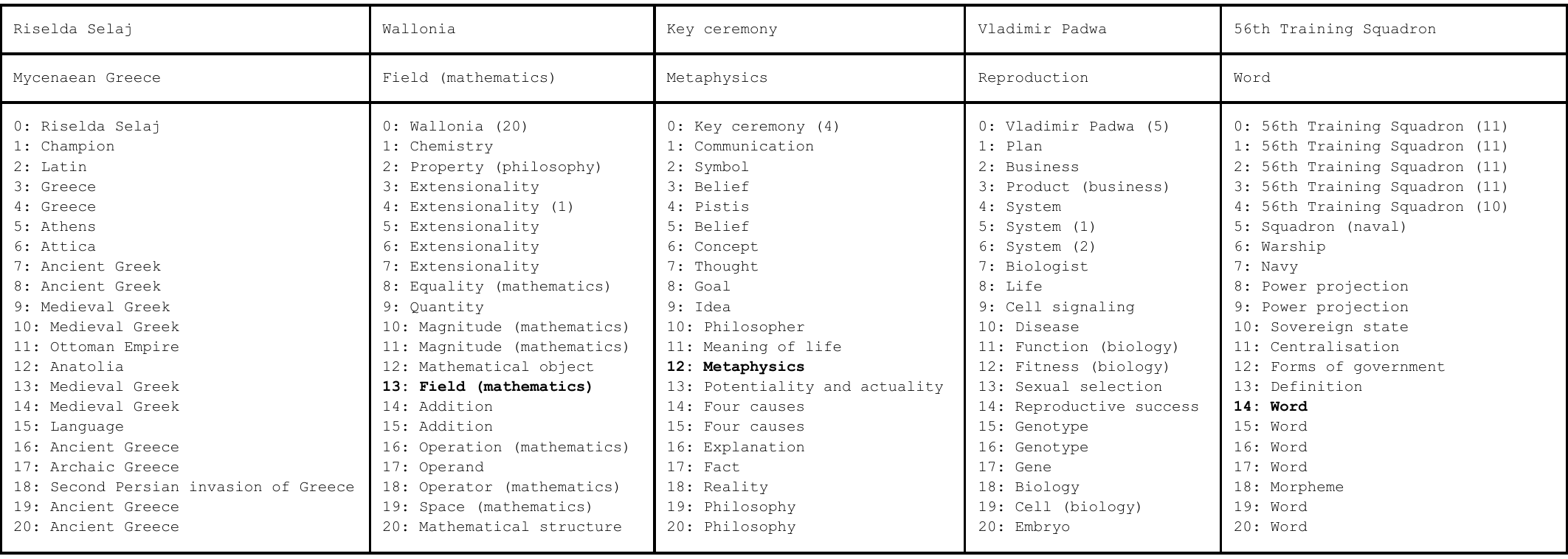}
    \vspace{-4mm}
    \caption{Example navigation trajectories in the 38M node Wikipedia graph. The start and target nodes are shown in the first two rows. Parentheses indicate paragraph block (zero-indexed) within article. Note that in cases where the agent fails to find the target node (cols 1 \& 3), it visits ones that are very close by.}
    \label{fig:traces}
    \vspace{-2mm}
\end{figure}

Table \ref{tab:fullnavigation} shows the results on the full graph, evaluating on $\{5,10,20\}$ step tasks, for both navigation and sentence search. Figure \ref{fig:traces} shows example trajectories of the trained agent. The first thing to note is that the overall navigation and sentence search performance of our methods is high. The 2018 Wikipedia graph we use for evaluation contains over $5$ million articles, over $38$ million nodes and $387$ million edges. On this graph (which we generalize to from the previous year's graph) our best method can find an article $20$ steps away $92.2\%$ percent of the time and can locate it given just a single sentence of that article $90.2\%$ of the time. This compares favorably with a similar evaluation performed in \citet{Nogueira16}, where their WebNav approach achieved a $12.5\%$ success rate for 16 step tasks on a 12M node graph. Compared to the performance of our method on the smaller $200$k graph, we see that performance has greatly improved, most likely due to the addition of orders of magnitude more training data. In particular we see that sentence search performance on $20$-step improves greatly from $34.6\%$ to a best performance of $90.2\%$.

Additionally, in all cases, our learned embeddings perform better than using the fixed features, derived from large language models. In general, feed-forward policies perform best, except for 20 step navigation tasks. For longer trajectories, the Transformer policy network achieves slightly better performance, while on sentence search, the feed-forward model is consistently superior. %

\begin{table}
  \vspace{-5mm}
  \caption{Full graph navigation (38M nodes) - Success rate (\%)}
  \centering
  \begin{tabular}{@{}lcccccc@{}}
    \toprule
    \multicolumn{1}{c}{} &
    \multicolumn{3}{c}{Navigation} & 
    \multicolumn{3}{c}{Sentence Search}
    \\
    \cmidrule(r){2-4}
    \cmidrule(r){5-7}
    Embedding + Policy     & 5 & 10 & 20 & 5 & 10 & 20 \\
    \midrule
    RoBERTA  + Feed-forward & 85.5  & 80.1 & 71.5 & 91.2 & 88.9 & 77.5  \\
    RoBERTA  + Transformer & 85.3 & 88.3 & 87.9 & 81.6 & 75.4 & 70.9  \\
    Embed train (ours)  + Feed-forward & 96.1  & 94.1 & 89.8 & 96.3 & 92.8 & 90.2 \\
    Embed train (ours)  + Transformer & 93.5  & 90.6 & 92.2 & 93.9 & 86.3 & 79.7 \\
    \bottomrule
  \end{tabular}
  \label{tab:fullnavigation}
  \vspace{-4mm}
\end{table}

\subsection{Application to fact verification}

\begin{table}[b!]
    \vspace{-5mm}
    \caption{Results on the FEVER benchmark (evidence retrieval only). The first two rows show the effect of adding our navigation scheme to the simple BM25 retrieval and TF-IDF re-ranker combination. Switching to a more powerful re-ranker (BigBird) results in a significant boost in F1 score, surpassing ~\citet{stammbach-2021} who also use BigBird~\citep{zaheer2020bigbird}.}
    \label{tab:fever}
    \centering
    \vspace{2mm}
    \begin{tabular}{@{}lccc@{}}
        \toprule
        Method & Precision@5 & Recall@5 & F1@5 \\
        \midrule
        BM25 + TF-IDF~\citep{fever}               & 0.33 & 0.40 & 0.36 \\
        BM25 + RFBC (Ours) + TF-IDF  & 0.38 & 0.55 & 0.46 \\
        BM25 + RFBC (Ours) + BigBird & 0.71 & 0.75 & 0.73 \\
        \citet{stammbach-2021}        & 0.26 & 0.94 & 0.41 \\
        \bottomrule
    \end{tabular}
\end{table}

\begin{figure}[b!]
    \centering
    \includegraphics[width=1.0\linewidth]{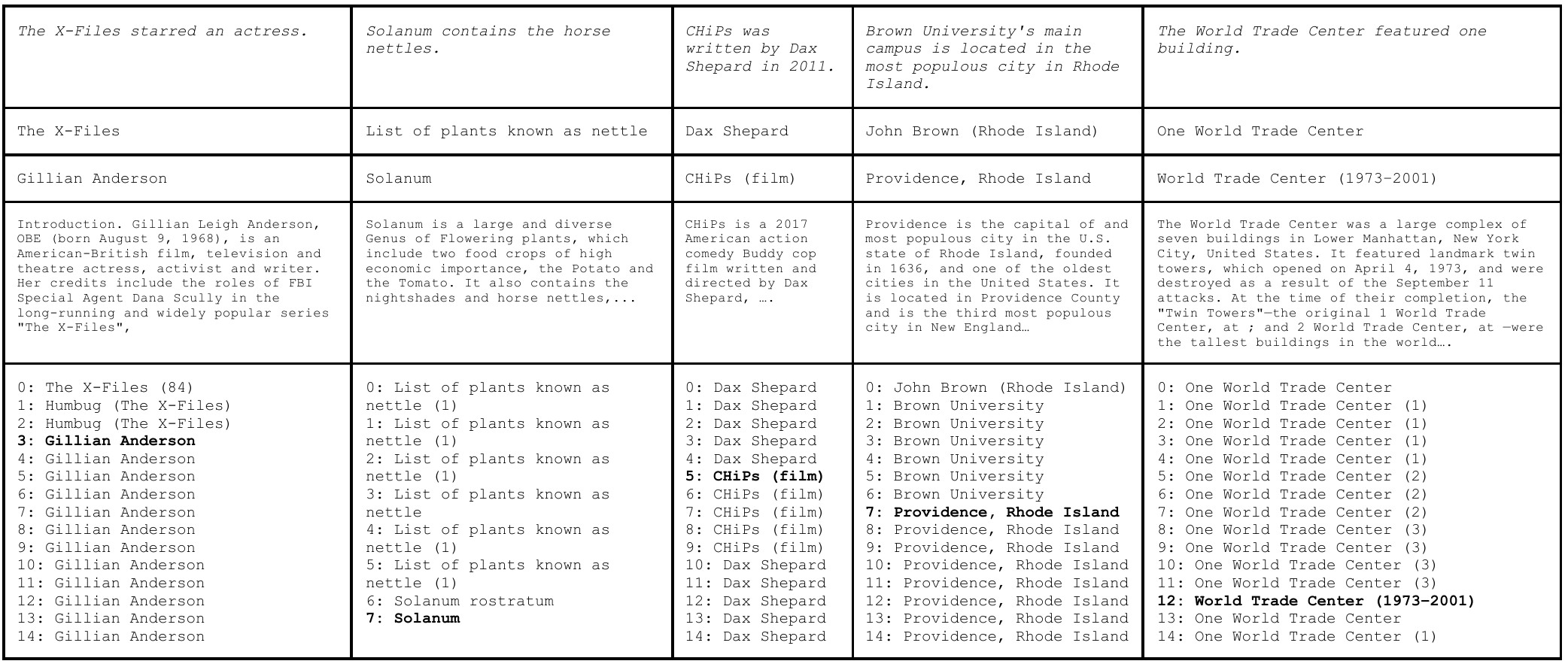}
    \vspace{-4mm}
    \caption{Successful navigation trajectories on the FEVER \citep{fever} evaluation set. Rows from top: claim to be verified; starting node (from BM25); ground truth target (not visible to agent); target node text; agent trajectory.}
    \label{fig:traces_fever}
    \vspace{-5mm}
\end{figure}

We now evaluate the ability of our navigation approach to select evidence on the FEVER development set \citep{fever} with results shown in Table~\ref{tab:fever}. The primary metric used is F1@top5 but we also give precision and recall. Paired with a basic retrieval (BM25) and re-ranker (TF-IDF), the approach obtains a respectable F1 of 0.46. If we remove our navigation component then this drops substantially to 0.36, demonstrating its utility. Swapping to a more powerful re-ranker (BigBird) boosts the F1 to 0.731, which is competitive with the state-of-the-art (FEVER \citet{leaderboard} rank \#6; rank \#1 F1 is 0.799). We also compare to a leading approach \citep{stammbach-2021} that similarly relies on the BigBird re-ranker, with our approach having a significantly better F1 score. A possible explanation is that their approach overwhelms the re-ranker with all possible hyperlinks whereas, navigation being inherently sparse, our model presents a more refined set for re-ranking. 
We note that our approach: (a) is significantly simpler than many top ranked approaches and (b) selects evidence over a much larger set than the curated version of Wikipedia used in FEVER (38M vs 5M). Examples of successful navigation traces are shown in Figure \ref{fig:traces_fever}.

\subsection{Application to question answering}
Finally, we present results on another downstream task where navigation helps.
In particular, we consider the task of finding correct evidence passage for open domain question answering.
We use the Natural Question (NQ) open-domain dataset presented in~\citet{orqa}.
Since we target navigating to the exact evidence passage required to answer the question, we use recall@\{1,2,3,4,5\} for finding the gold evidence passage as our metric.
More commonly the metric marks a retrieved passage to be correct if it contains the answer string, but this causes a lot of false positives, e.g. the answer string appears in a totally irrelevant context.

\begin{table}[ht]
    \vspace{-3mm}
    \centering
    \caption{Results on the Natural Questions open domain benchmark (evidence retrieval only). Third and fourth rows show the effect of adding our navigation scheme to the simple BM25 retrieval and BigBird re-ranker.
    First row is a reference re-training of state-of-the-art method on our setup.}
    \label{tab:nq-res}
    \scalebox{0.93}{
    \begin{tabular}{@{}lccccc@{}}
    \toprule
    & Recall@1 & Recall@2 & Recall@3 & Recall@4 & Recall@5 \\
    \midrule
    RocketQA~\citep{qu2021rocketqa} & 32.7 & 43.8 & 51.7 & 58.3 &  62.5 \\
    \midrule
    BM25 &  11.1 & 17.3 & 21.8 & 24.9 & 27.3 \\
    BM25 + BigBird ReRank & 22.2 & 30.2 & 35.1 & 38.7 &  41.2 \\
    \midrule
    BM25 + RFBC (Ours) + BigBird ReRank & 31.6 & 41.3 & 46.6 & 49.8 & 51.6 \\
    Improvement & +42.3\% & +36.7\% & +32.7\% & +28.7\% & +25.2\% \\
    \bottomrule
    \end{tabular}
    }
\end{table}

The results are tabulated in Table~\ref{tab:nq-res}.
Paired with a basic retrieval (BM25) and re-ranker (BigBird), the approach obtains a respectable Recall@1 of 31.6.
If we remove our navigation component then this drops substantially to 22.2, demonstrating its utility.  
As a reference, we also ran a state-of-the-art system RocketQA~\citep{qu2021rocketqa} on our setup and evaluated it using our harder metric.
It is worthwhile to note that our approach (a) is significantly simpler than state-of-the-art approaches like RocketQA which employ 4 stages of dual-encoder and cross-attention models, and (b) selects evidence over a larger set than the commonly used preprocess Wikipedia passages (38M vs 21M).

\section{Discussion}
We have presented a simple and effective scheme for navigating a large Wikipedia graph that is applicable to more general web navigation problems. We show that behavioral cloning of random trajectories is a viable approach to learning both entity embeddings and a navigation policy. When applied to the fact verification task on FEVER dataset and the open-domain question answering task on NQ dataset, they offer highly competitive performance, whilst being complementary to existing approaches.
Another advantage worth highlighting is that the navigating agent provides a provenance on how it arrived at the relevant evidence, which many other methods (like dense passage retrieval) do not provide.
One limitation of our approach when we move from Wikipedia to the wider Internet is that our scheme relies on a good target encoder. For navigation and our downstream tasks, there was a clear ground-truth target node available training, but in other settings this might not be the case. 
Another limitation is that we require a re-ranker to decide on the final retrieved sentence, but ideally the agent would decide for itself when it has reached the correct node. %

\bibliographystyle{abbrvnat}
\bibliography{main}

\newpage

\appendix

\section{Data / Environment details}
\label{sec:wiki_env}

\subsection{Wikipedia data processing}
\label{sec:appendix_data}

We start by downloading English Wikipedia snapshot (the pages-articles.xml.bz2 file) from Wikimedia\footnote{\url{https://dumps.wikimedia.org/enwiki/}} or Internet Archive.\footnote{\url{https://archive.org/search.php?query=Wikimedia\%20database\%20dump\%20of\%20the\%20English\%20Wikipedia}}
We extract text from English Wikipedia for a given snapshot using Gensim's WikiCorpus class.\footnote{\url{https://github.com/RaRe-Technologies/gensim/blob/master/gensim/corpora/wikicorpus.py}}
For each page, this tool extracts the plain text and hyperlinks, and strips out all structured data such as lists and figures.
To avoid pages without sufficient textual content, we filter out categories, listical, disambiguation pages and any other page with less than 200 characters. 
Each article ($\mu=1000$ words) is split into paragraph-sized blocks ($\mu=100$ words), which form the \emph{nodes} in our Wikipedia Graph. 
This was done both for conceptual reasons (humans don't read an entire article at once but look at bits and pieces of it at one go) and for modeling reasons (most language models have an upper input token length that is smaller than the average Wikipedia page). 
The \emph{edges} in the graph are formed in three ways:
\begin{enumerate}
    \item for nodes in the same article, we add ``previous node'' and ``next node'' links to connect them in a chain; 
    \item organic hyperlinks -- the internal links connecting Wikipedia pages to each other; and
    \item we additionally run entity linking\footnote{We use a standard linker from \url{https://cloud.google.com/natural-language/docs/analyzing-entities}} because repeat mentions of an entity in Wikipedia lack hyperlinks, which is problematic when articles are split up. 
\end{enumerate}
This constitutes our Wikipedia graph construction, where the text blocks act as vertices and the organic hyperlinks along with entity links are the edges. 
We store the graph as an adjacency list with metadata, using a memory mapped key-value data-structure to enable fast random access during navigation.

We run the above graph generation pipeline for two different snapshots of Wikipedia:
\begin{enumerate}
    \item June 01 2017\footnote{\url{https://archive.org/download/enwiki-20170601/enwiki-20170601-pages-articles.xml.bz2}}: This corresponds to FEVER fact verification dataset. 
    \item December 2018\footnote{\url{https://archive.org/download/enwiki-20181220/enwiki-20181220-pages-articles.xml.bz2}}: This corresponds to Natural Questions open dataset.
\end{enumerate}

\subsection{Smaller 200k node graph construction}
\label{sec:small_graph}
We construct smaller train and eval graphs with 200k nodes each for the following reasons:
\vspace{-1mm}
\begin{enumerate}[topsep=1pt,itemsep=1pt,partopsep=1pt, parsep=1pt]
    \item Checking strict generalization by making train and eval graphs to be totally disjoint, i.e. no common nodes or edges. The full 2017 and 2018 are sufficiently different but not entirely disjoint.
    \item Running experiments on full graphs are expensive, so for more thorough evaluation across multiple baselines we constructed the smaller graphs.
\end{enumerate}

To construct these two smaller graphs, we start with the full 2018 graph. We sort the nodes by their in-degrees.
We mark nodes with odd ranks to be in train and even ranks to be in eval, which ensures complete separation between the two.
For the train graph we start with rank 1 node (and for eval with rank 2 node) and then select all its neighbors which have odd rank (even rank).
Then all the odd (even) nodes connected to the selected ones are picked.
This process is continued until the desired number of nodes are selected.

\subsection{Graph statistics}
\label{sec:wiki_stats}
We construct two versions of the full Wikipedia graph following the method outlined in Appendix~\ref{sec:appendix_data}.
The main summary statistics (number of articles, nodes, edges, words per node, median path length) of the resulting graphs are listed in Table~\ref{tab:graphsize}. 

\begin{figure}[ht]
    \centering
    \hfill
    \begin{subfigure}{0.33\linewidth}
        \includegraphics[width=\textwidth]{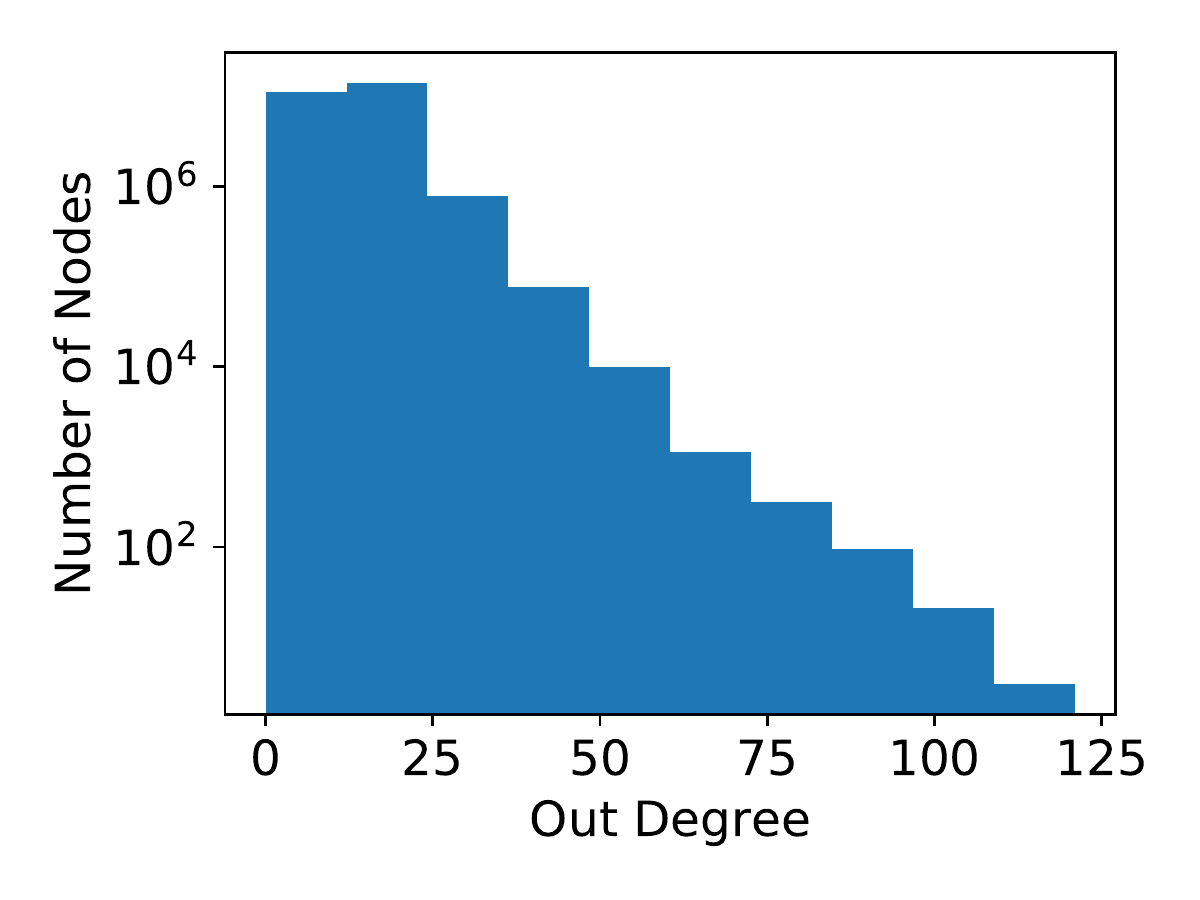}
        \caption{2017: Out-degree}
        \label{fig:deg_dist_2017}
    \end{subfigure}
    \hfill
    \begin{subfigure}{0.33\linewidth}
        \includegraphics[width=\textwidth]{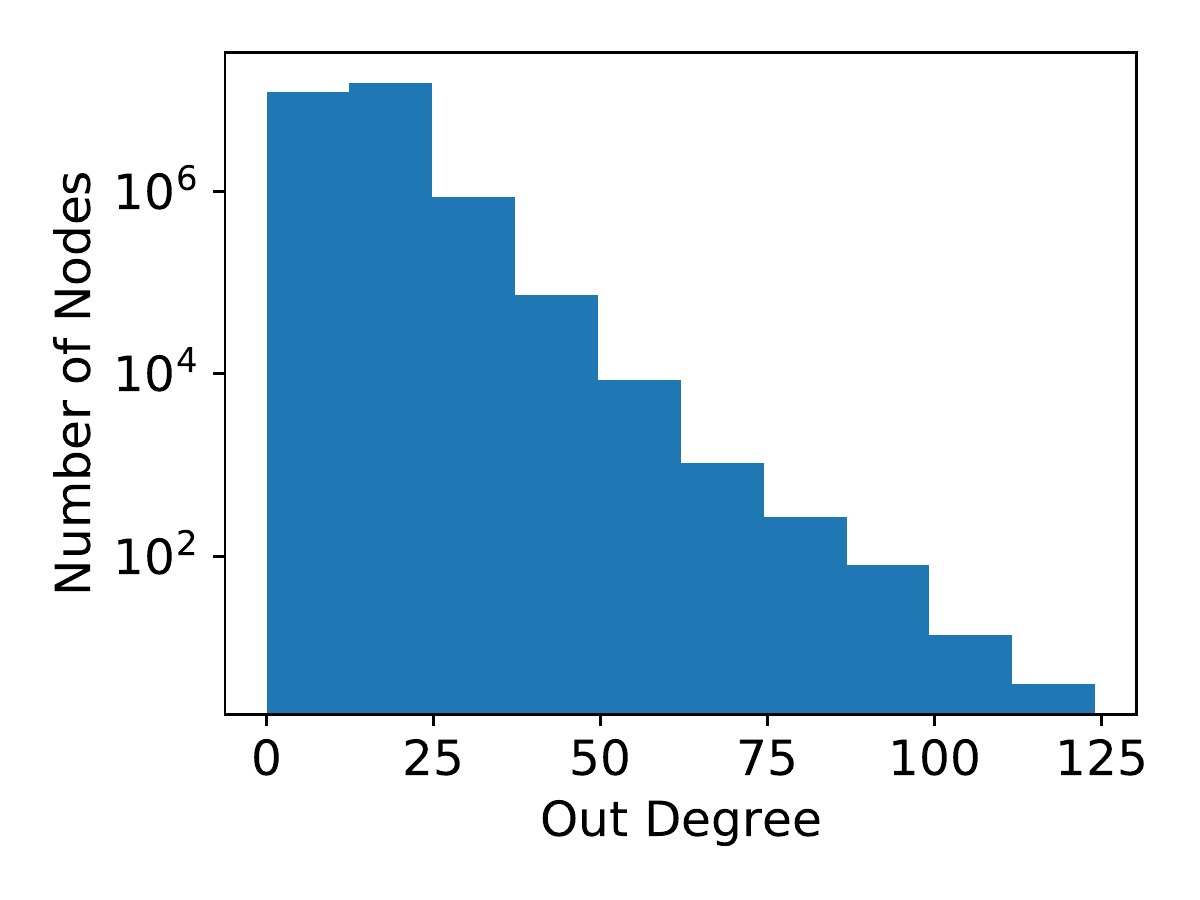}
        \caption{2018: Out-degree}
        \label{fig:deg_dist_2018}
    \end{subfigure}
    \hfill
    \phantom{}\\
    \hfill
    \begin{subfigure}{0.33\linewidth}
        \includegraphics[width=\textwidth]{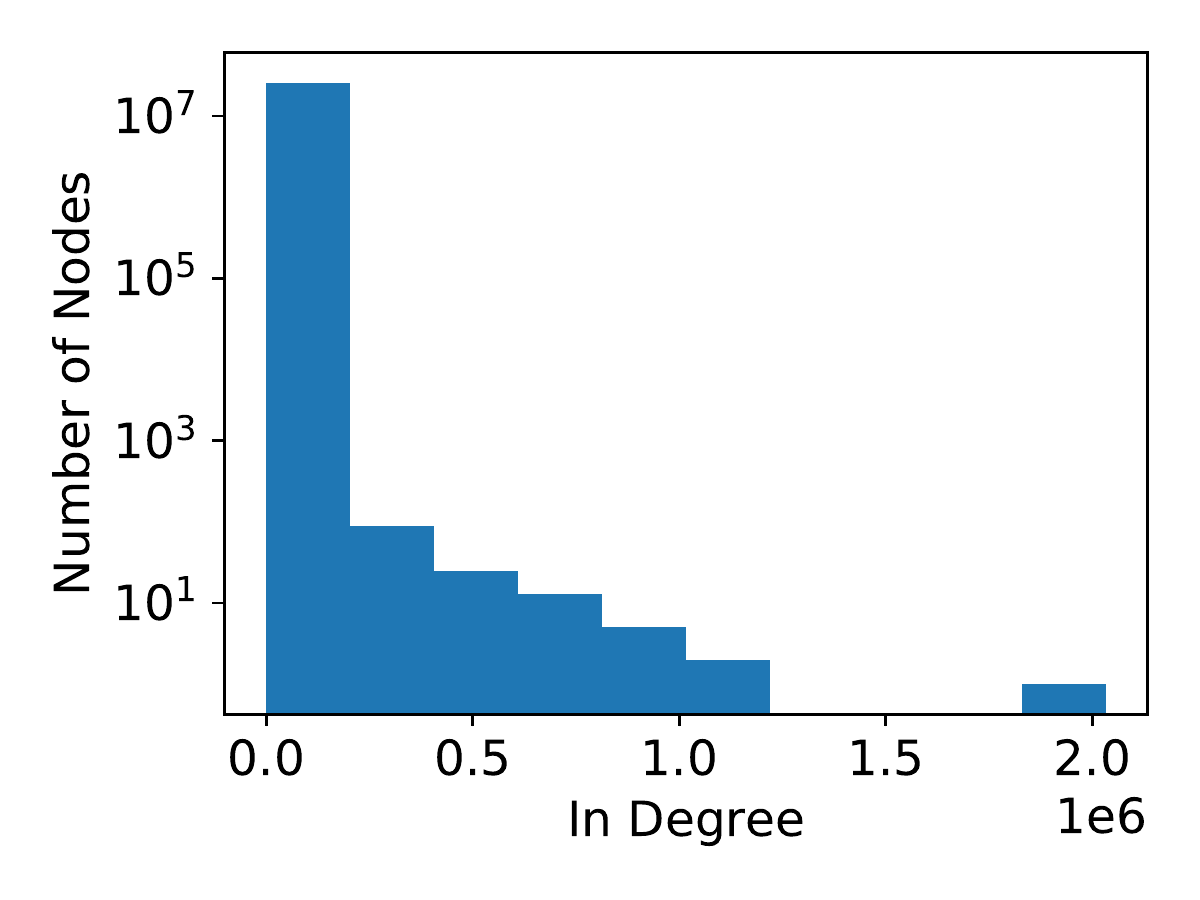}
        \caption{2017: In-degree}
        \label{fig:in_deg_dist_2017}
    \end{subfigure}
    \hfill
    \begin{subfigure}{0.33\linewidth}
        \includegraphics[width=\textwidth]{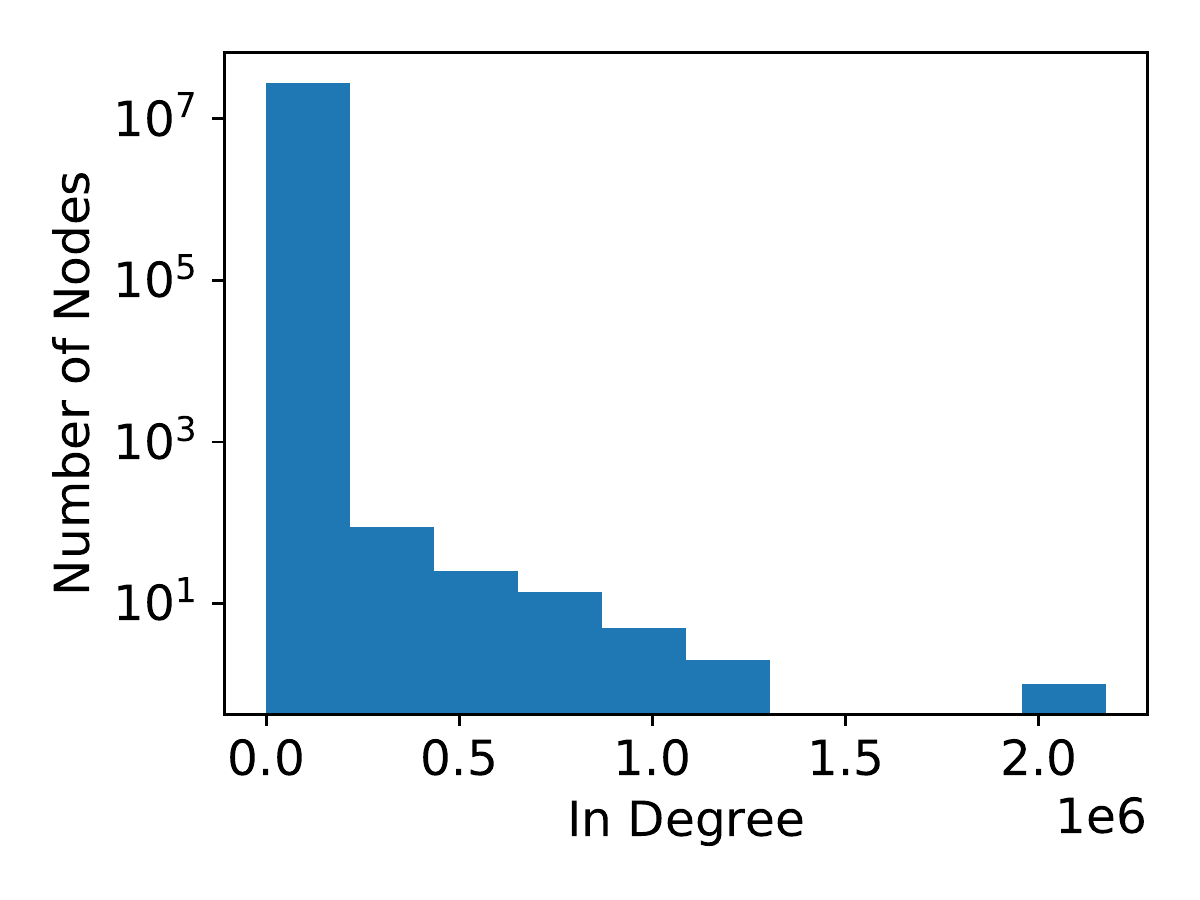}
        \caption{2018: In-degree}
        \label{fig:in_deg_dist_2018}
    \end{subfigure}
    \hfill
    \vspace{-1mm}
    \caption{Degree Distribution}
    \label{fig:deg_dist}
\end{figure}

A more detailed picture of the graph can be obtained by looking at its degree distribution, which is presented in Figure~\ref{fig:deg_dist}.
It shows the graph has tell-tale sign of web-graph: it has a power-law distribution with a few hub nodes.

\begin{figure}[ht]
    \centering
    \hfill
    \begin{subfigure}{0.33\linewidth}
        \includegraphics[width=\textwidth]{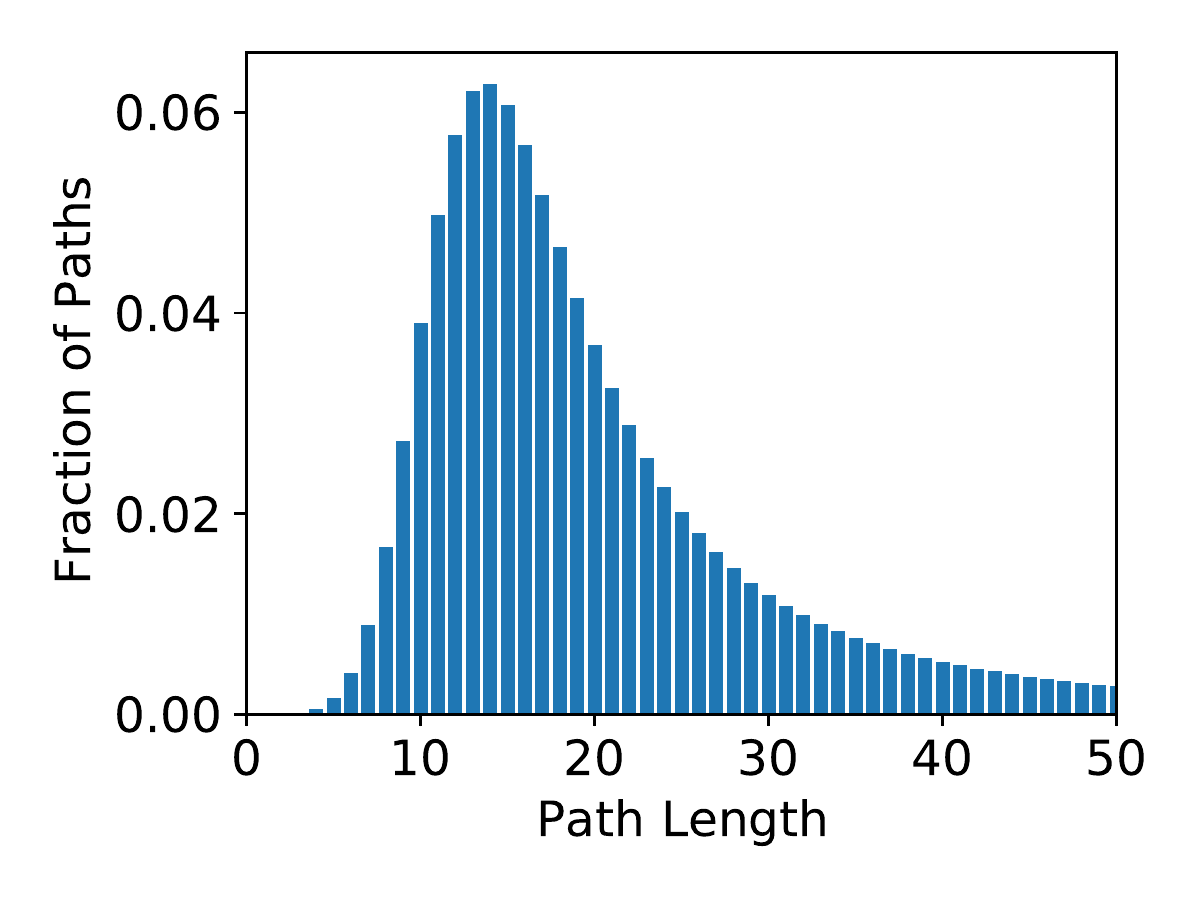}
        \caption{PDF}
        \label{fig:dspl-pdf-zoom}
    \end{subfigure}
    \hfill
    \begin{subfigure}{0.33\linewidth}
        \includegraphics[width=\textwidth]{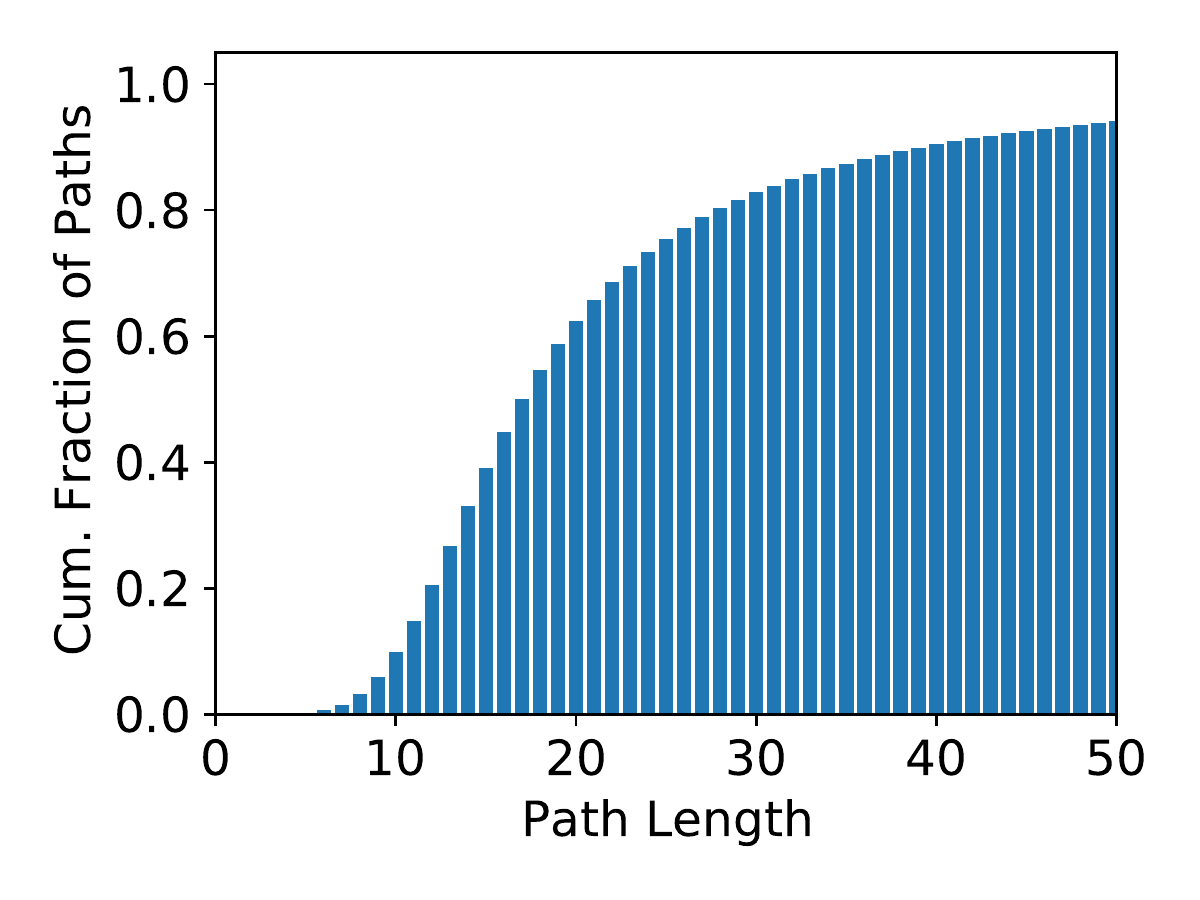}
        \caption{CDF}
        \label{fig:dspl-cdf-zoom}
    \end{subfigure}
    \hfill
    \phantom{}\\
    \hfill
    \begin{subfigure}{0.33\linewidth}
        \includegraphics[width=\textwidth]{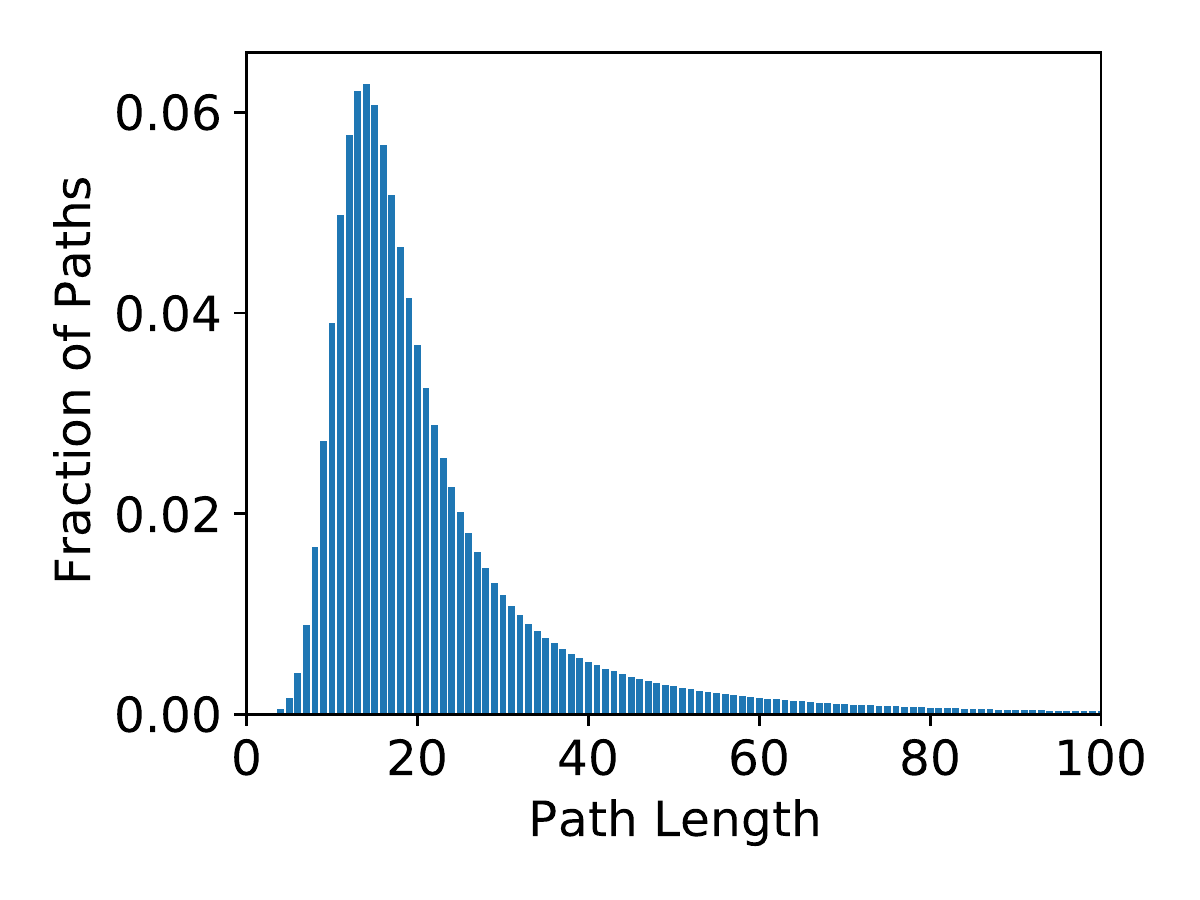}
        \caption{PDF}
        \label{fig:dspl-pdf}
    \end{subfigure}
    \hfill
    \begin{subfigure}{0.33\linewidth}
        \includegraphics[width=\textwidth]{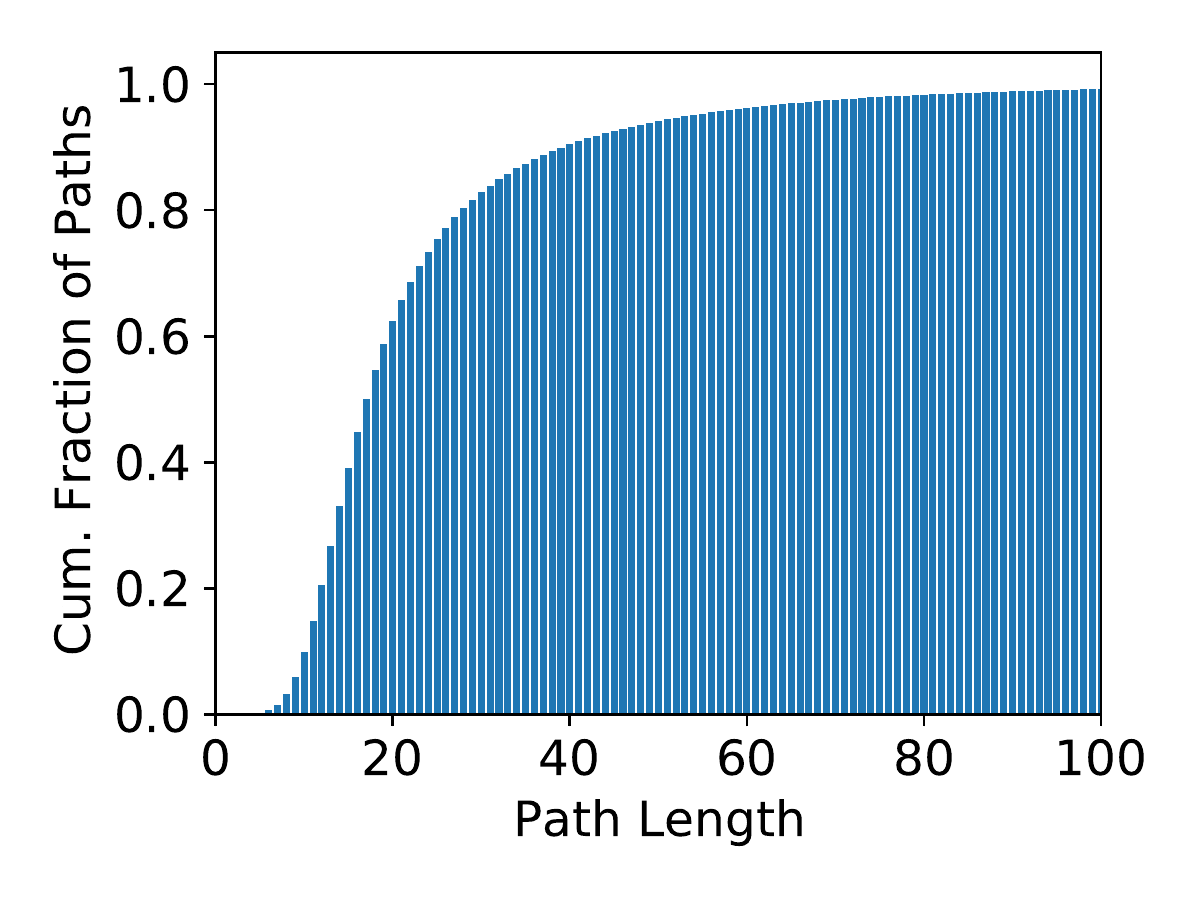}
        \caption{CDF}
        \label{fig:dspl-cdf}
    \end{subfigure}
    \hfill
    \vspace{-1mm}
    \caption{Distribution of Short Path Length estimated by computing shortest path between 1 trillion pairs of random nodes in the 2018 graph.}
    \label{fig:dspl}
\end{figure}

We try to estimate how far are two nodes in the graph.
Calculating the entire distribution of shortest path length is prohibitively expensive for such large graph.
We follow~\citet{ye2010distance} method to get an approximation.
We start by selecting 250k nodes having highest in-degree, which is a good approximation of the giant component in a web graph as required by ~\citet{ye2010distance}.
We then perform a single source to all shortest paths on each of these 250k nodes using Dijkstra's algorithm for sparse graphs.
This resulted in computing approximately 1 trillion shortest paths based on which the estimated distribution of shortest path length is shown in Figure~\ref{fig:dspl}.
 
\begin{wrapfigure}{r}{0.4\textwidth}
  \vspace{-5mm}
  \centering
  \includegraphics[width=0.33\textwidth]{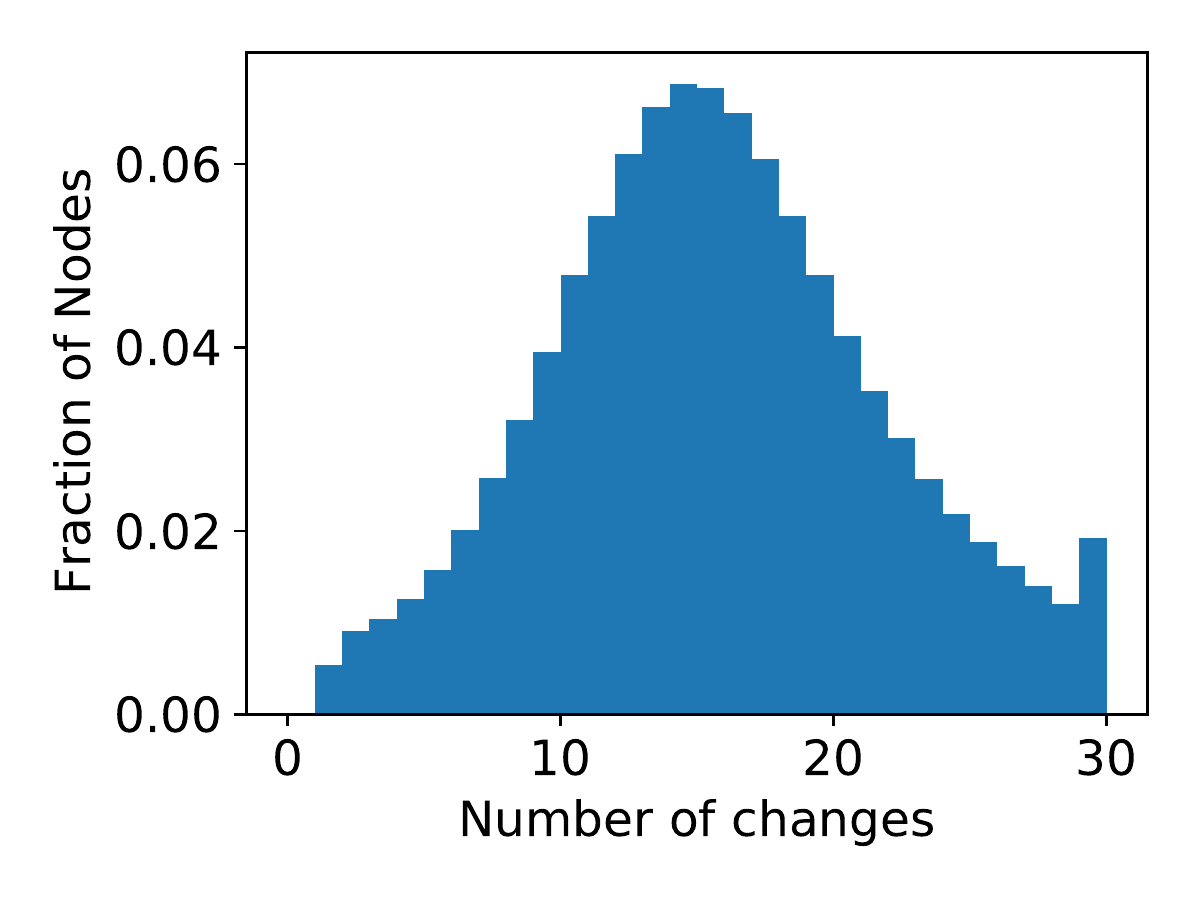}
  \vspace{-3mm}
  \caption{Number changes in edges among common nodes between 2017 and 2018}
  \label{fig:diff-graph}
  \vspace{-9mm}
\end{wrapfigure}
Finally, we would like to point out that there is significant evolution (difference) between the 2017 and 2018 graph.
Some important statistics reflecting the changes across the two graphs are:
\vspace{-1mm}
\begin{itemize}[topsep=1pt,itemsep=1pt,partopsep=1pt, parsep=1pt]
    \item New articles: 404,773 (7.5\%)
    \item New nodes: 3,871,578 (10\%)
    \item Deleted articles: 58,476 (1.1\%)
    \item Deleted nodes: 874,372 (2.3\%)
\end{itemize}
Further in Figure~\ref{fig:diff-graph}, we show histogram of changes in edges among the common nodes between the two graphs.
Apart from addition/deletion of organic hyperlinks, just modifications of text would yield to differences as the chunking will be different.
This shows significant generalization is needed to successfully navigate across the two graphs, simply memorization will not work.

\section{Experimental details}
\subsection{Embedding pretraining}
\label{sec:will_pretrain}
This section details our embedding pretraining procedure, as introduced in Section~\ref{sec:approach}.
\paragraph{Model details}
The transformer is fine-tuned from a pre-trained RoBERTa \citep{Roberta} model. As in Equation \ref{eq:bc} our model is trained to optimize
\begin{align}
    \mathcal{L}(\theta) = \mathbf{E}_{\mathcal{E}(n_0, \ldots, n_T)}\left[ \sum_{t=1}^{T-1}\log p_\theta(n_t | n_{t}, n_g)  \right]
    \label{eq:bc_app}
\end{align}
where we set $n_g = n_T$. 
We parameterize this distribution as
\begin{align}
    p_\theta(n_{t+1}|n_t, n_g) \propto \exp(f_\theta(s_\phi(n_g), s_\phi(n_t), a_\phi(n_t, n_{t+1})))
\end{align}
where $s_\phi(\cdot)$ and $a_\phi(\cdot, \cdot)$ are functions which extract embeddings for an entire node and a node-action combination, respectively and $f_\theta(\cdot, \cdot, \cdot)$ is a function which combines these embeddings to produce action probabilities. At their core, $s_\phi$ and $a_\phi$ are based on the same transformer model. We tokenize the text at node $n$ to produce $L$ tokens and pass these tokens through the transformer to produce $L$ embeddings.

\begin{figure}[ht!]
    \centering
    \includegraphics[width=0.85\linewidth]{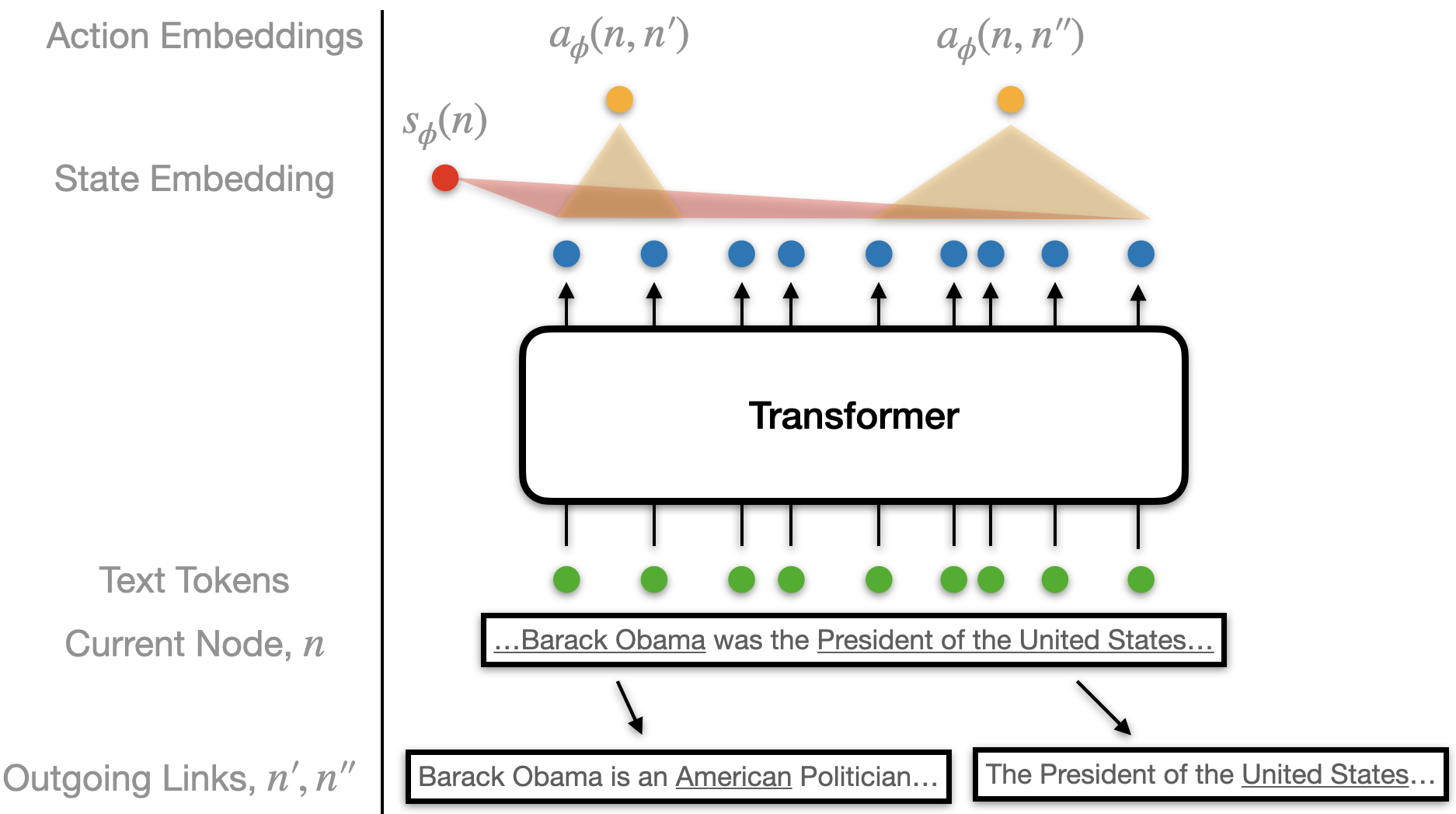}
    \caption{Description of how embeddings are extracted for navigation pretraining.}
    \label{fig:app}
\end{figure}

The state representation $s_\phi(n)$ is produced by taking the mean of these embeddings and passing this vector through a $\texttt{tanh}$ nonlinearity. The node-action embedding $a_\phi(n, n')$ embeds the action of moving from node $n$ to its neighbor $n'$. Let us assume the text at node $n$ is ``\underline{Barack Obama} was the \underline{President of the United States} during...'' where the substrings ``Barack Obama'' and ``President of the United States'' correspond to hyperlinks to other neighboring articles $n'$ and $n''$. Then we construct the node-action embedding $a_\phi(n, n')$ by passing the tokenized text of node $n$ through our transformer, again producing $L$ transformed token embeddings. Then we take the transformed tokens which correspond to the text ``Barack Obama'' and take their mean and pass this through the $\texttt{tanh}$ nonlinearity. If we instead wanted to produce $a_\phi(n, n'')$ we would take the transformed tokens which correspond to the text ``President of the United States'' instead, take the mean and apply the nonlinearity. This is visualized in Figure \ref{fig:app}. 

The function $f_\theta$ which combines these embeddings first concatenates them and passes the combined input through a simple neural network with a single hidden layer and 1 output neuron.

\paragraph{Training details}
We train these models to optimize Equation \ref{eq:bc_app}. The objective is optimized with the Adam~\citep{kingma2014adam} optimizer using an initial learning rate of $10^{-5}$ with a one-cycle cosine decay schedule and linear warm up of $10$K steps. The training batch size is $512$ and the model is trained over $3$M iterations. Node text is clipped (or padded) to $200$ tokens.
The weights of the transformer are initialized from a pretrained RoBERTa~\citep{Roberta} masked language model.

\subsection{Small graph experiments}
\label{app:smallgraph}
\paragraph{Model details}
In the small graph experiments, we use the simple policy network $FF$ as described in Section~\ref{sec:model}. As discussed, we use a 1-layer MLP (meaning a single learned linear layer with no activation function). We use a hidden size of $768$ to match the size of $\phi$. Because the final logits are the result of an inner product between each state and action embedding of size $768$, to aid network training, we add a normalization layer (jax.nn.normalize) before the inner product.

For the action embedding $a_i = \phi(s_i)$ we also concatenate a 1-hot vector that represents the action type (e.g. whether it was a link action, or a next action, or a prev action). We also concatenate another bit that indicates whether the state $s_i$ of the particular action has been visited before. 

For the purposes of the RL baseline, we can view graph navigation as a goal-conditioned MDP.\footnote{On a known fixed graph, we can formulate this an an MDP. However, in the case where the graph changes or in a setting where there are multiple possible graphs it is trivial to reformulate this as a POMDP where we only can see the current node and neighboring nodes} Our observation is the same as the inputs to the RFBC models. The reward function $R$ for current node $n$ and goal node $n_g$ we can write simply as:
\begin{align}
R = \left \{ \begin{array}{ll}
        1 & n = n_g  \\
        0 & otherwise \\
    \end{array} \right.
\end{align}
For the RL baseline, we used IMPALA~\citep{Espeholt2018IMPALASD}, which is a kind of policy gradient approach, and thus has a policy network $\pi$ and a ``baseline'' or value function network. Since the policy network has the same output space as our RFBC models, we use an identical architecture for this. For the value network, we use a very similar architecture. We again use $\phi$ as our encoding of the current node and the target node, again concatenate them together. We then feed this through a 1-layer MLP to compute the value.

\paragraph{Training details}
The procedure for generating the trajectories we use for the BC loss is described in Section~\ref{sec:trajdist}. Except for our experiment in Section~\ref{sec:investtrajectory}, we always use the random forward trajectories. Following \citet{victoria}, in the navigation experiments, we randomly drop out edges in the training graph with probability $0.5$ to reduce over-fitting. As stated in Section~\ref{sec:tasks}, we set max steps to $B=100$. For RFBC training on the small graph, we use RMSProp with a learning rate of $0.01$, a decay of $0.9$ an epsilon of $10^{-10}$. 

For the sentence search tasks, our models train a separate $\phi$ for the target embedding. As we stated, we use MiniBERT for this embedding $\phi_\text{target}$ and train it with the same optimization settings, except that we reduce the learning rate for these weights to $10^{-4}$. We use a batch size of $512$ and we train for $50,000$ network update steps.

For RL training, we use VTRACE~\citep{Espeholt2018IMPALASD} loss, using the reward above. We again use RMSProp with a learning rate of $0.01$, a decay of $0.9$ an epsilon of $10^{-10}$. We set the baseline cost to $0.5$, trajectory length (the number of timesteps the RL loss backprops through) to $100$, batch size again to $512$ and max update steps to $50,000$. We did a parameter sweep for entropy costs of $0.1, 0.01, 0.001$ and gamma of $0.8, 0.9$. To give RL the best chance possible, we choose the maximum over this sweep for each experiment (but still none of these match RFBC performance).

\subsection{Full Wikipedia navigation experiments}
\label{sec:full_wiki_nav}
On the full Wikipedia training experiments in Section~\ref{sec:fullwikiexp}, we evaluate on navigation and sentence search using either our feed-forward or transformer model and using either RoBERTA fixed embeddings, or our trained embeddings described in Appendix~\ref{sec:will_pretrain}. 

The feed-forward model is identical to the one described in Appendix~\ref{app:smallgraph}. The transformer model is the standard transformer model from \citet{vaswani2017attention} with $4$ layers, an attention size of $64$, $12$ heads, and mlp hidden size of $3072$ and dropout rate of $0$. DistillBERT~\citep{Sanh2019DistilBERTAD} is used for $\phi(n_g)$ for the sentence search experiments. 

The learning parameters and all other relevant training parameters for all models are identical to those in Appendix~\ref{app:smallgraph} except that the learning rate for the transformer model is lower ($10^{-4}$) as the earlier learning rate was too high for transformer models.

\subsection{RFBC + RL}

\begin{figure}[h!]
    \centering
    \includegraphics[width=0.5\linewidth]{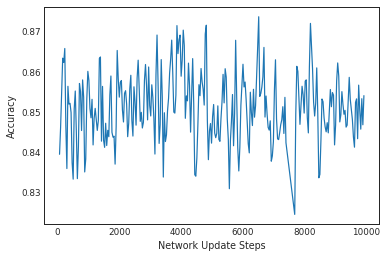}
    \caption{Training curve for RFBC with RL finetuning.}
    \label{fig:RFBC+RL}
\end{figure}

We additionally try finetuning an RL policy starting from our RFBC-trained navigation policies. We use the same network and learning settings as we did in the other RL experiments. We train for an additional 10,000 network updates (512M environment steps). We add the final numbers in Table~\ref{tab:simplenavigation} and show the training curve in Figure~\ref{fig:RFBC+RL}. We can see from the training curve that RL training accuracy fluctuates around the point where RFBC training left off. From the results in Table~\ref{tab:simplenavigation} we see little statistically significant difference (about 1 point in either direction).

\section{Downstream task details}

\subsection{FEVER experiments}
\label{sec:fever_eval}
FEVER~\citep{fever} is a dataset containing $185,445$ claims labeled as \textsc{Supported}, \textsc{Refuted}, or \textsc{NotEnoughInfo}. Verifiable claims (i.e. \textsc{Supported} or \textsc{Refuted}) are annotated with evidence sentences supporting this classification, which are drawn from a preprocessed version of the June 2017 Wikipedia snapshot. Notably, only the article introductions are retained, and claims are generated from a curated set of approximately $50,000$ popular pages.

\paragraph{Data processing}
To make an aligned navigation graph for FEVER, we start with the June 01 2017 snapshot of Wikipedia and build the graph using the procedure described in Section~\ref{sec:appendix_data}. Because of differences in text preprocessing, FEVER's version of Wikipedia does not precisely align to our own. We then try to match FEVER's Wikipedia articles to our graph, relying on Wikipedia
URL redirections for handling disambiguations. This resulted in $99.5\%$ matching of articles between our navigation graph and the FEVER version of Wikipedia. Second, we align the evidence sentences in FEVER to sentences from our text blocks using a fuzzy string match. Specifically, we use the token set ratio to score the similarity between an evidence sentence and the text in a graph node. If the score between a sentence and a node is $\geq 80$, the node is added as an evidence node for the given claim. This threshold is sufficiently high to minimize the chances of a false positive, while also matching a high percentage of evidence sentences. Some sentences are difficult to match, particularly those that are split between two text blocks. Ultimately, $93.1\%$ of all evidence sentences were matched to a node. This gives us an augmented FEVER dataset, where for each claim we have a corresponding set of evidence nodes in the graph.
These nodes are then used as navigation targets for fine-tuning during the training phase. We generate trajectories for BC by running BM25 over all nodes in the graph, taking the top 10 matches as starting nodes, and finding shortest paths to evidence nodes.

\begin{figure}[t]
    \centering
    \hfill
    \begin{subfigure}{0.33\linewidth}
        \includegraphics[width=\textwidth]{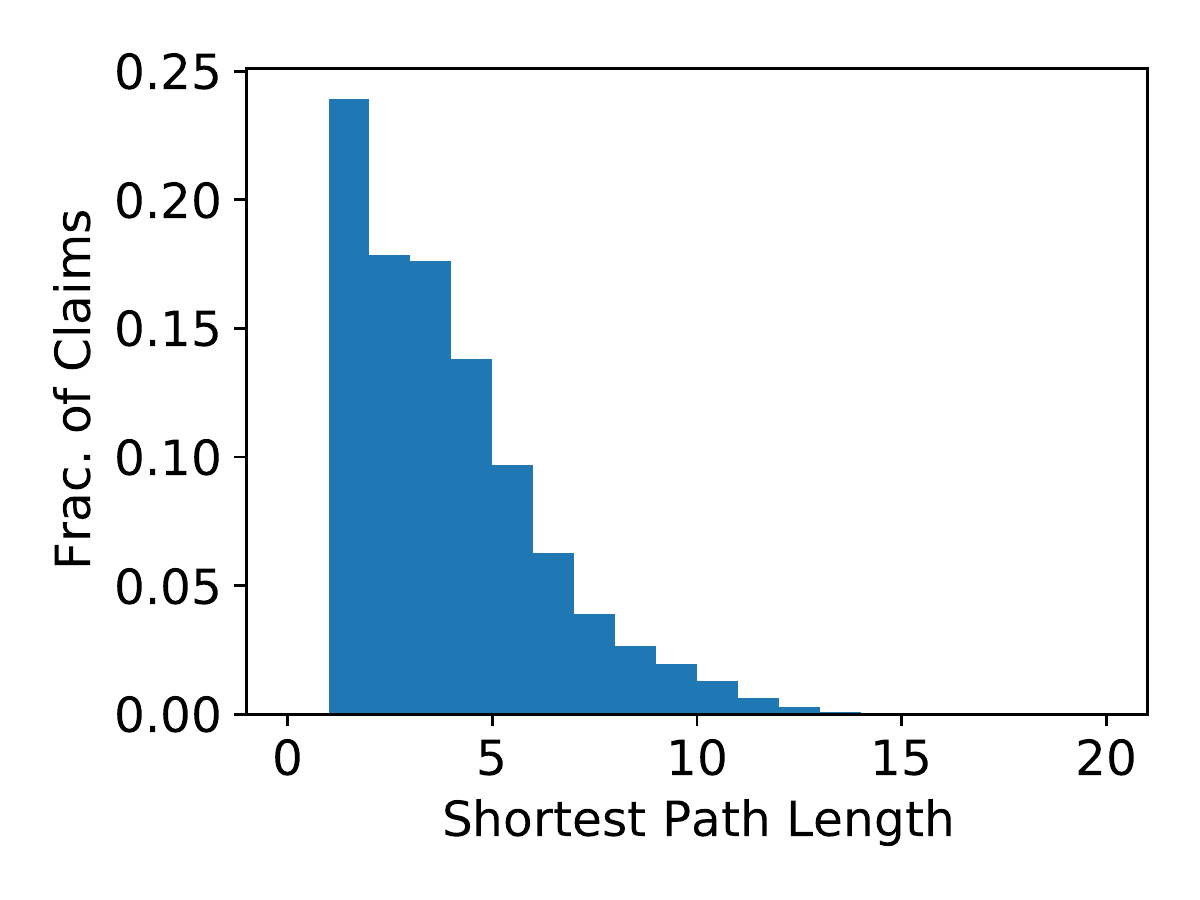}
        \caption{Train/Density}
        \label{fig:fever-dspl-train-pdf}
    \end{subfigure}
    \hfill
    \begin{subfigure}{0.33\linewidth}
        \includegraphics[width=\textwidth]{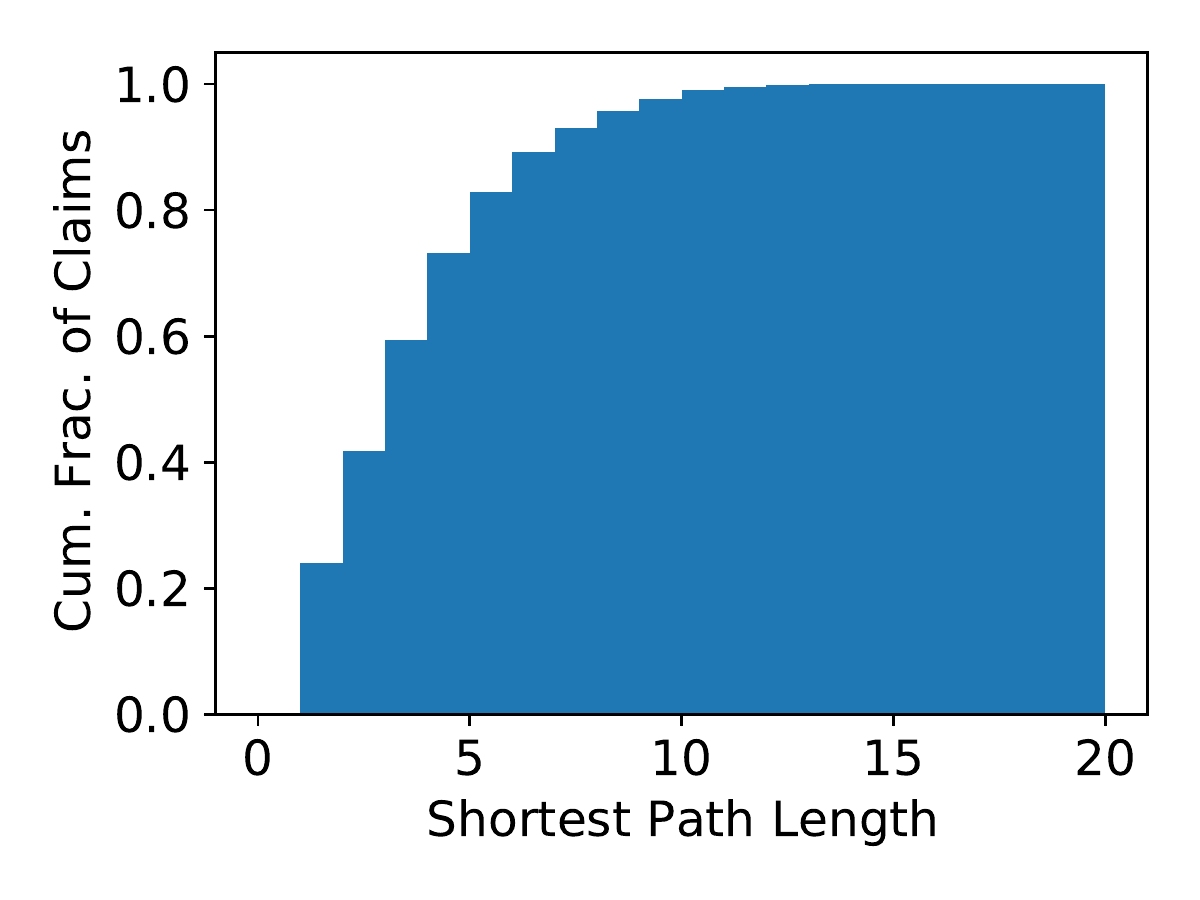}
        \caption{Train/Cumulative}
        \label{fig:fever-dspl-train-cdf}
    \end{subfigure}
    \hfill
    \phantom{}\\
    \hfill
    \begin{subfigure}{0.33\linewidth}
        \includegraphics[width=\textwidth]{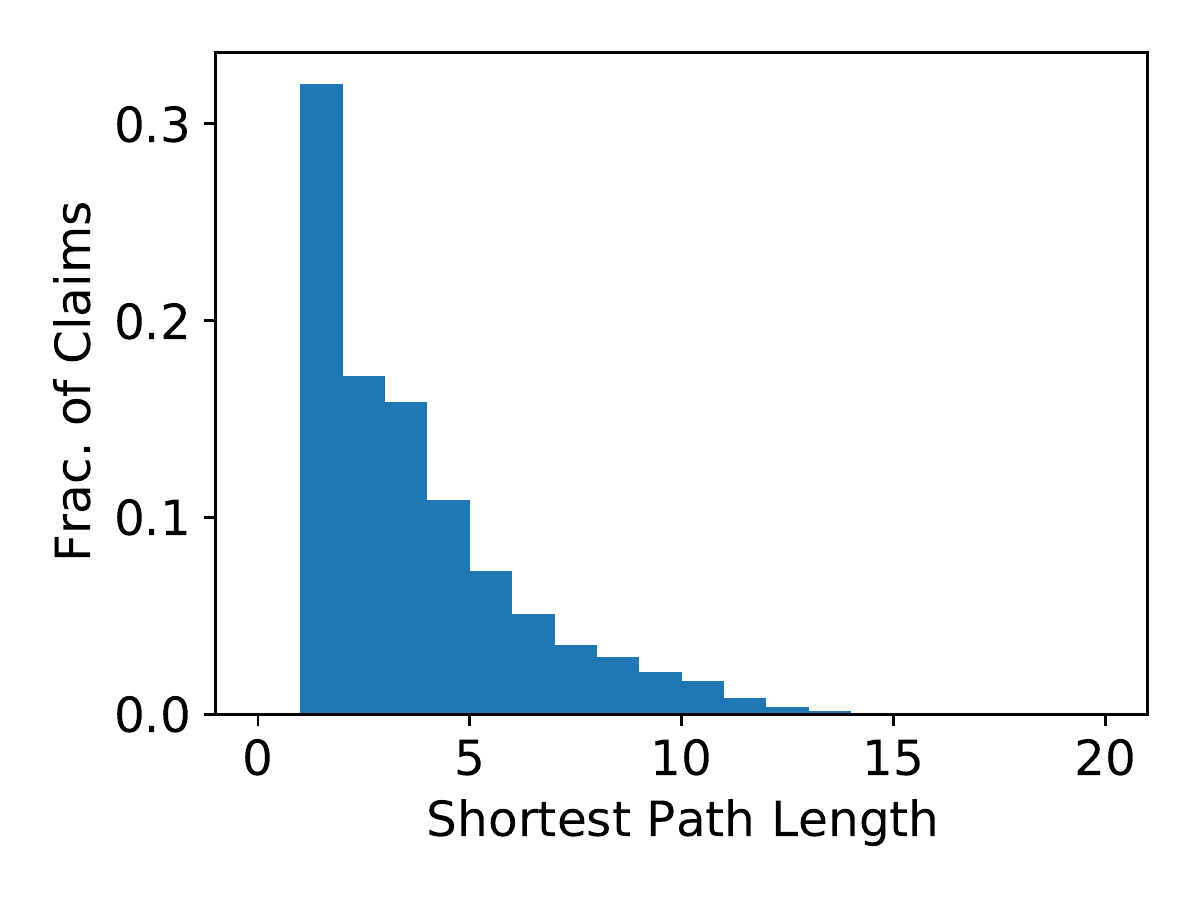}
        \caption{Test/Density}
        \label{fig:fever-dspl-test-pdf}
    \end{subfigure}
    \hfill
    \begin{subfigure}{0.33\linewidth}
        \includegraphics[width=\textwidth]{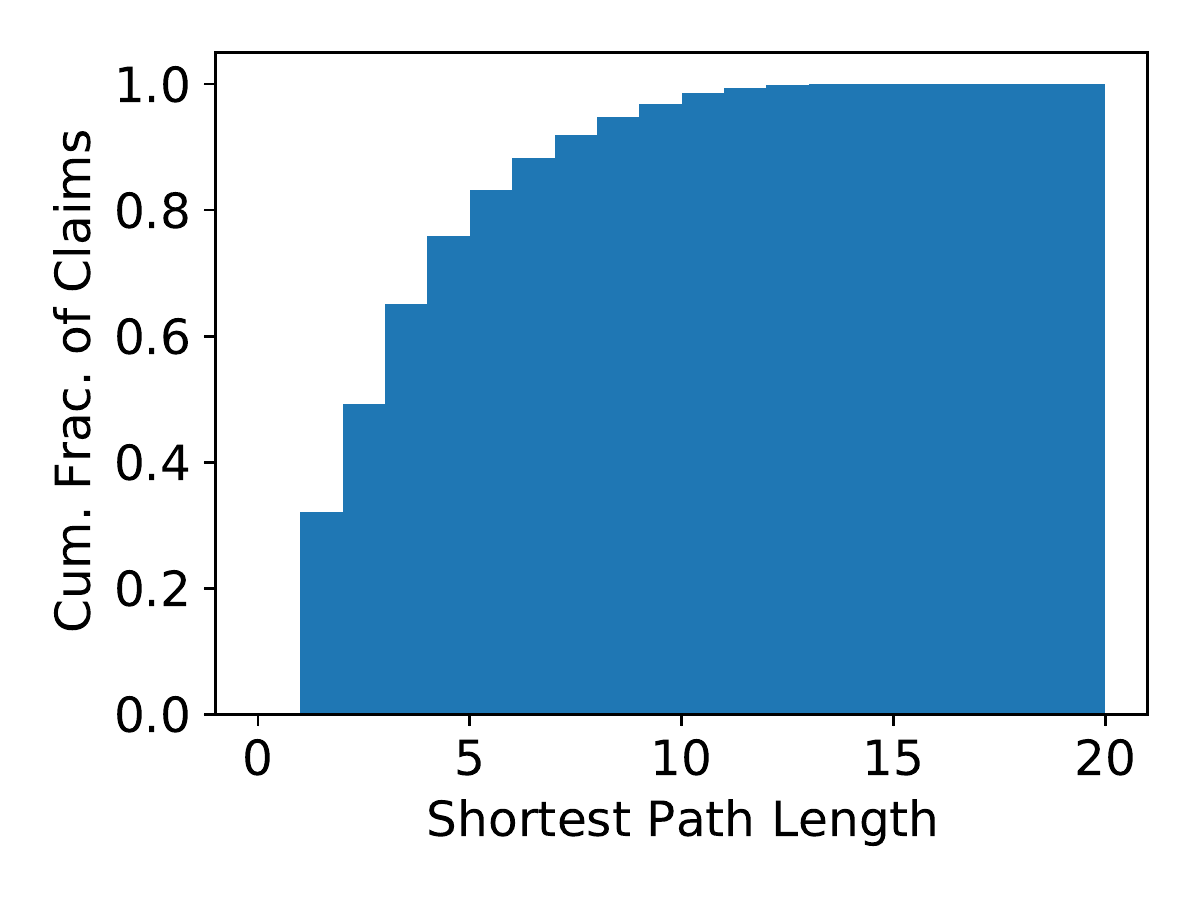}
        \caption{Test/Cumulative}
        \label{fig:fever-dspl-test-cdf}
    \end{subfigure}
    \hfill
    \vspace{-1mm}
    \caption{Distribution of Short Path Length from starting node computed by BM25 to target node on FEVER dataset on 2017 graph.}
    \label{fig:fever-dspl}
\end{figure}

\paragraph{Data statistics}
We compute the shortest-path distance between top-1 retrieval of BM25 and target node in this graph in Figure~\ref{fig:fever-dspl}.
As can be seen that most path lengths are relatively short, which implies that BM25 lands us in the right vicinity and by a small amount of navigation around we will be able to find the right evidence passage.

\paragraph{Training details}
For the FEVER benchmark, our model uses learned embeddings for $\phi$ with a single feed-forward layer for the policy network. The model is pretrained on the 
$5$-step sentence search task, and finetuned on trajectories generated from the augmented FEVER dataset using the the following loss function: \begin{equation}
    \mathcal{L}(\theta) = \mathcal{L}_{\rm BC}(\theta) + 0.1 \| \phi_\text{target}({\rm claim}) - \phi(s_g)   \|_2
\end{equation}
where $\mathcal{L}_{\rm BC}$ is the normal BC loss we use in our other experiments. 

Only the $\phi_\text{target}$ weights are finetuned for this task. We use AdamW with a constant learning rate of $10^{-6}, \beta_1=0.9, \beta_2=0.999$, and $\epsilon=10^{-6}$. The model is finetuned for $100,000$ update steps with a batch size of $512$. No edge dropout is used during finetuning.

\paragraph{Evaluation details}
For each claim in the FEVER development dataset, we first run BM25 over all nodes in the graph and take the top $5$ matches as starting nodes. From each node, the finetuned model then navigates for $20$ steps. We collect all sentences in all nodes visited by the agent over the 100 total navigation steps, and match them to FEVER evidence sentences using the WRatio function in the fuzzywuzzy package.\footnote{\url{https://pypi.org/project/fuzzywuzzy/}} Two sentence ranking methods are explored: Gensim's TF-IDF implementation,\footnote{\url{https://radimrehurek.com/gensim/models/tfidfmodel.html}} and the BigBird re-ranker of \citet{stammbach-2021}, using their open source model and weights.\footnote{\url{https://github.com/dominiksinsaarland/document-level-FEVER}} Finally, the top-5 evidence sentences are then submitted to the official FEVER scorer\footnote{\url{https://github.com/sheffieldnlp/fever-scorer/blob/9d9ed27637adddf73bc2f8e38c436bdc032c9f1f/src/fever/scorer.py}} to compute the accuracy, recall, and F1 scores.

\subsection{NQ experiments}
\label{sec:nq_eval}
Natural Questions (NQ) is a dataset collected from real users asking questions on the Google search engine which are answerable using Wikipedia.
We use the Natural Question open-domain subset presented in~\citet{orqa} which has been aligned to Wikipedia dump of December 20, 2018 by~\citet{karpukhin20dpr}.
In this split it has 58,880 questions for training and another 8,757 questions as development set for evaluation.
(The test set is not aligned to Wikipedia passages so we do not evaluate on it.)

\paragraph{Data processing}
Most prior works in literature utilize the processed collection of 21M passages provided by~\citet{karpukhin20dpr}, but unfortunately we cannot use it as it has no hyperlink information which is crucial for our graph building.
So we re-align the questions in NQ to our 2018 graph with 38M passages (i.e. nodes) which has different text blocks than~\citet{karpukhin20dpr}.
As the article title corresponding to the passage was given as part of the dataset, the alignment task was local to the document.
We could find all the article titles, i.e. 100\% match in locating the document in our dump.
As before, to align the evidence passage in NQ to our text blocks, we first used a fuzzy string match.
Then in the ranked list of fuzzy string matches, we searched for exact answer string.
The highest ranked node with answer string was labelled as the ground truth.
Some answer strings were difficult to match, particularly those that are split between two text blocks. Ultimately, $99.3\%$ of all evidence passages were matched to a node. 
This gives the re-aligned NQ dataset, where for each question we have a corresponding set of evidence nodes in our graph.
These nodes are then used as navigation targets for fine-tuning during the training phase. We generate trajectories for BC by running BM25 over all nodes in the graph, taking the top 10 matches as starting nodes, and finding shortest paths to evidence nodes.

\begin{figure}
    \centering
    \hfill
    \begin{subfigure}{0.33\linewidth}
        \includegraphics[width=\textwidth]{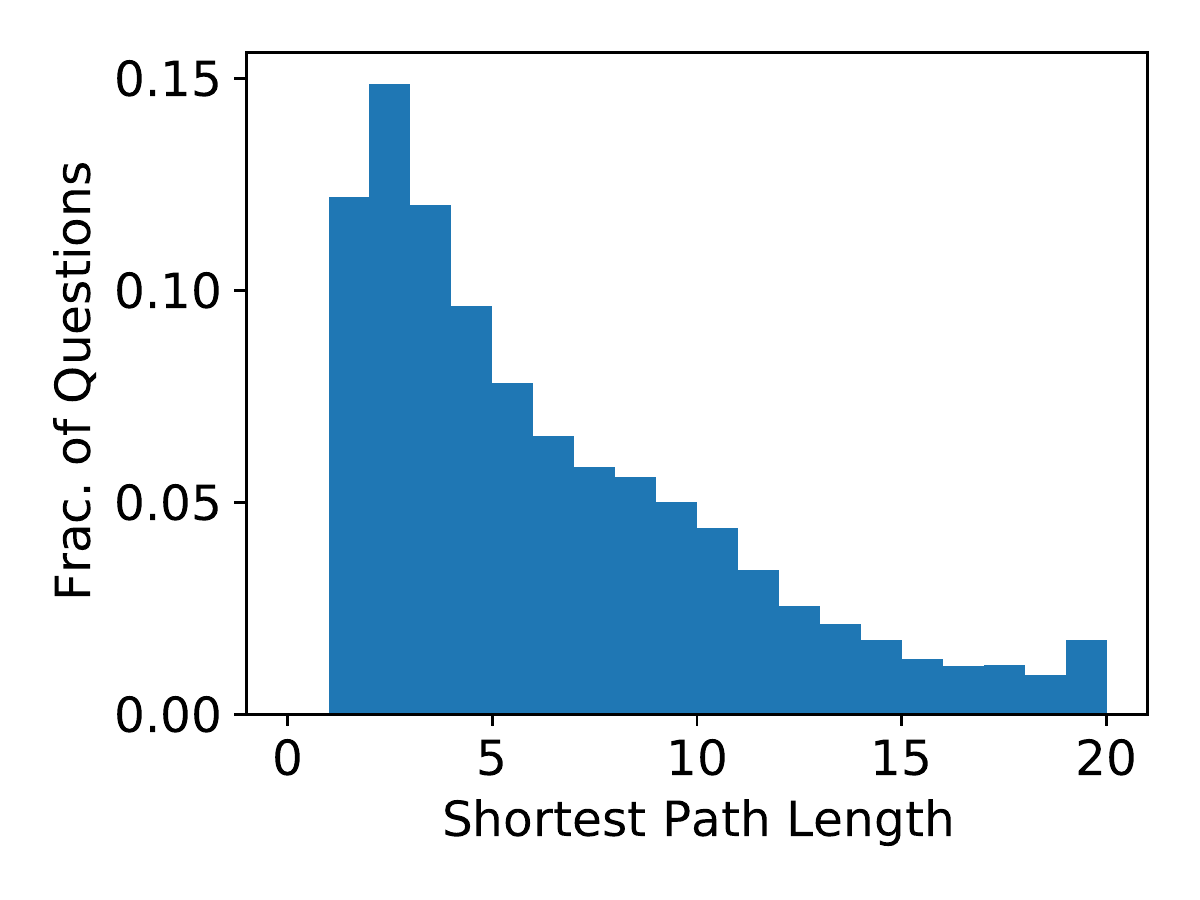}
        \caption{Train/PDF}
        \label{fig:nq-dspl-train-pdf-zoom}
    \end{subfigure}
    \hfill
    \begin{subfigure}{0.33\linewidth}
        \includegraphics[width=\textwidth]{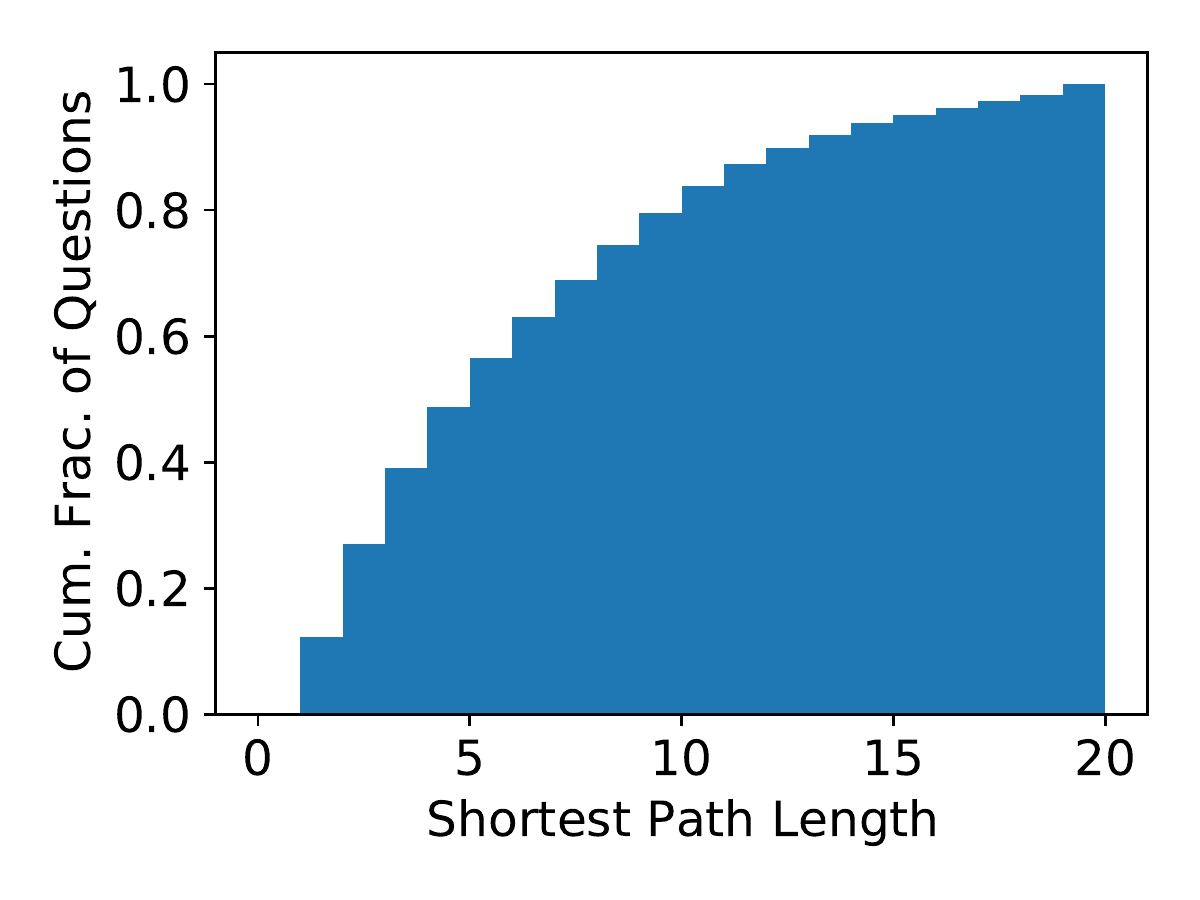}
        \caption{Train/CDF}
        \label{fig:nq-dspl-train-cdf-zoom}
    \end{subfigure}
    \hfill
    \phantom{}\\
    \hfill
    \begin{subfigure}{0.33\linewidth}
        \includegraphics[width=\textwidth]{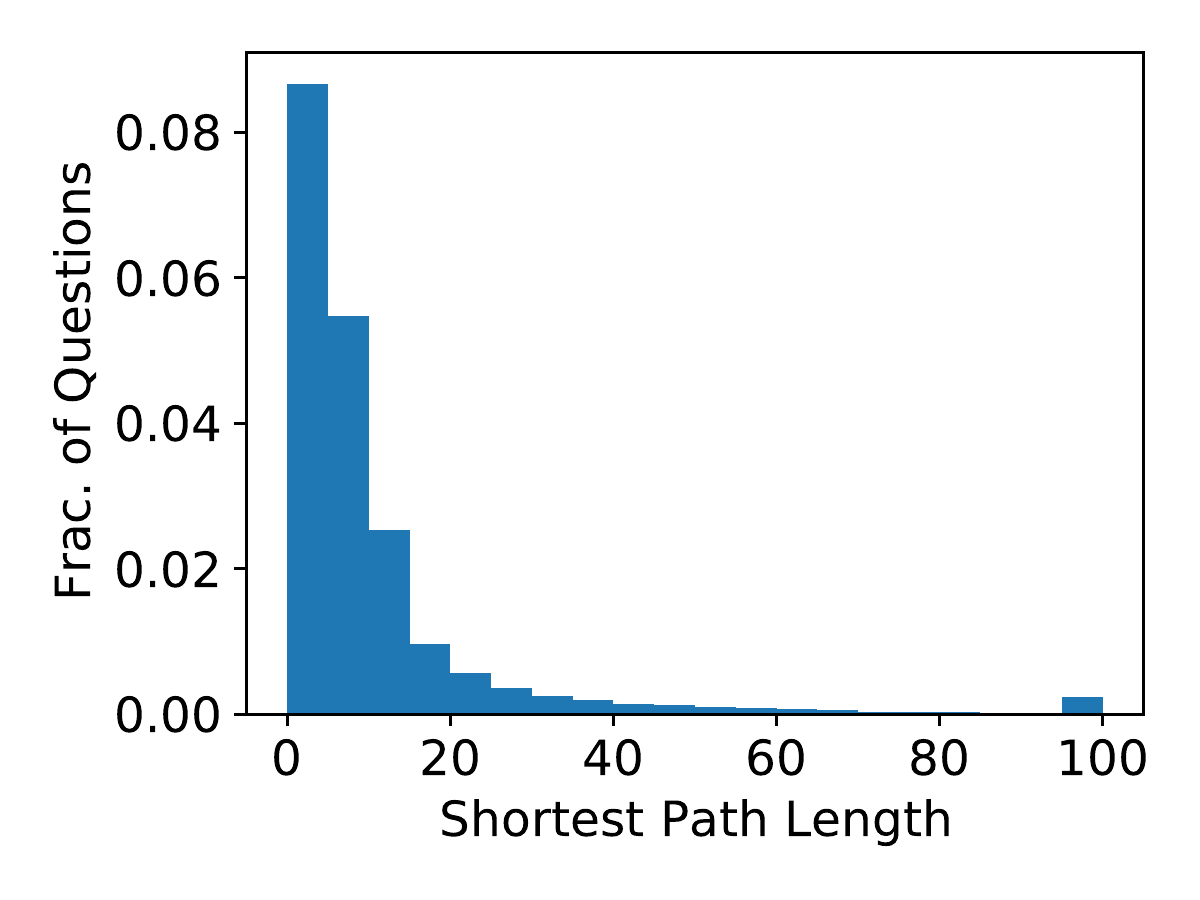}
        \caption{Train/PDF}
        \label{fig:nq-dspl-train-pdf}
    \end{subfigure}
    \hfill
    \begin{subfigure}{0.33\linewidth}
        \includegraphics[width=\textwidth]{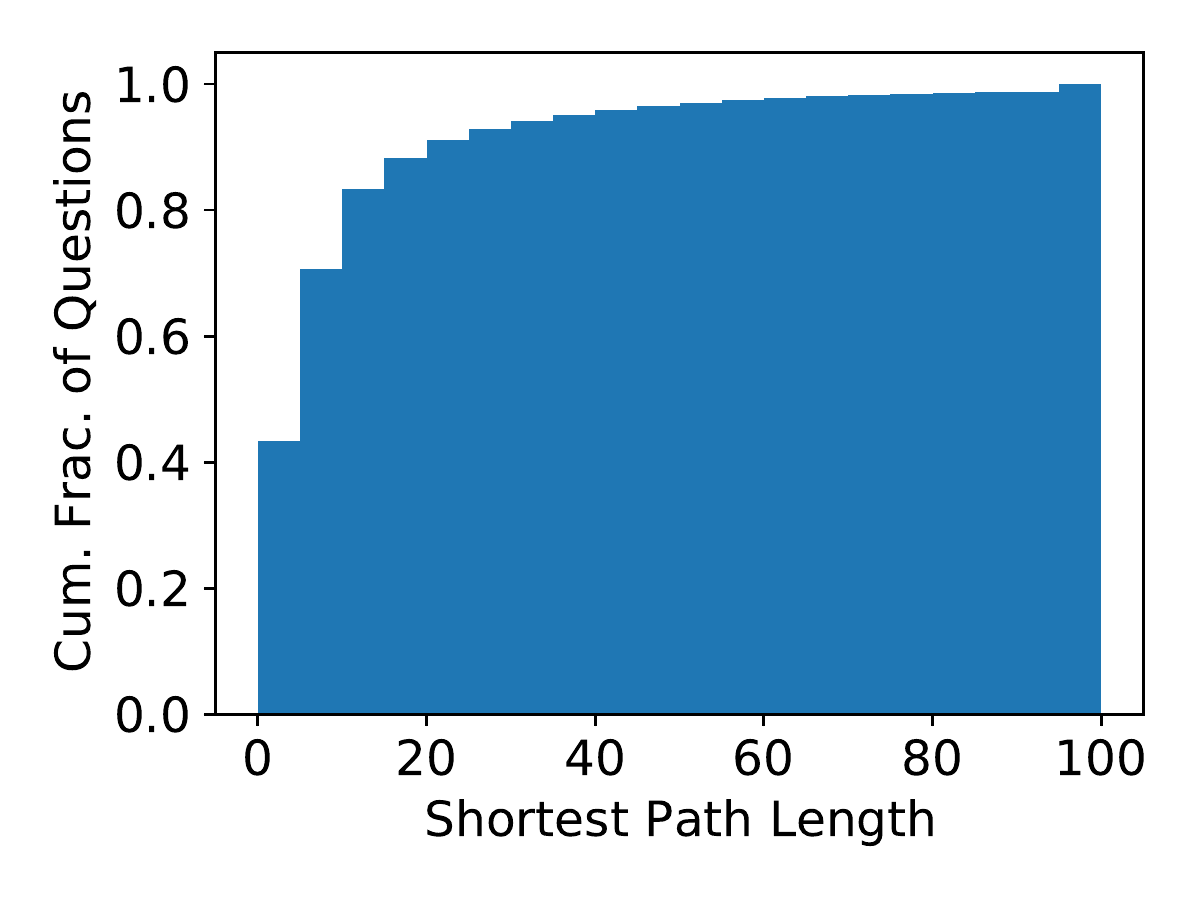}
        \caption{Train/CDF}
        \label{fig:nq-dspl-train-cdf}
    \end{subfigure}
    \hfill
    \phantom{}\\
    \hfill
    \begin{subfigure}{0.33\linewidth}
        \includegraphics[width=\textwidth]{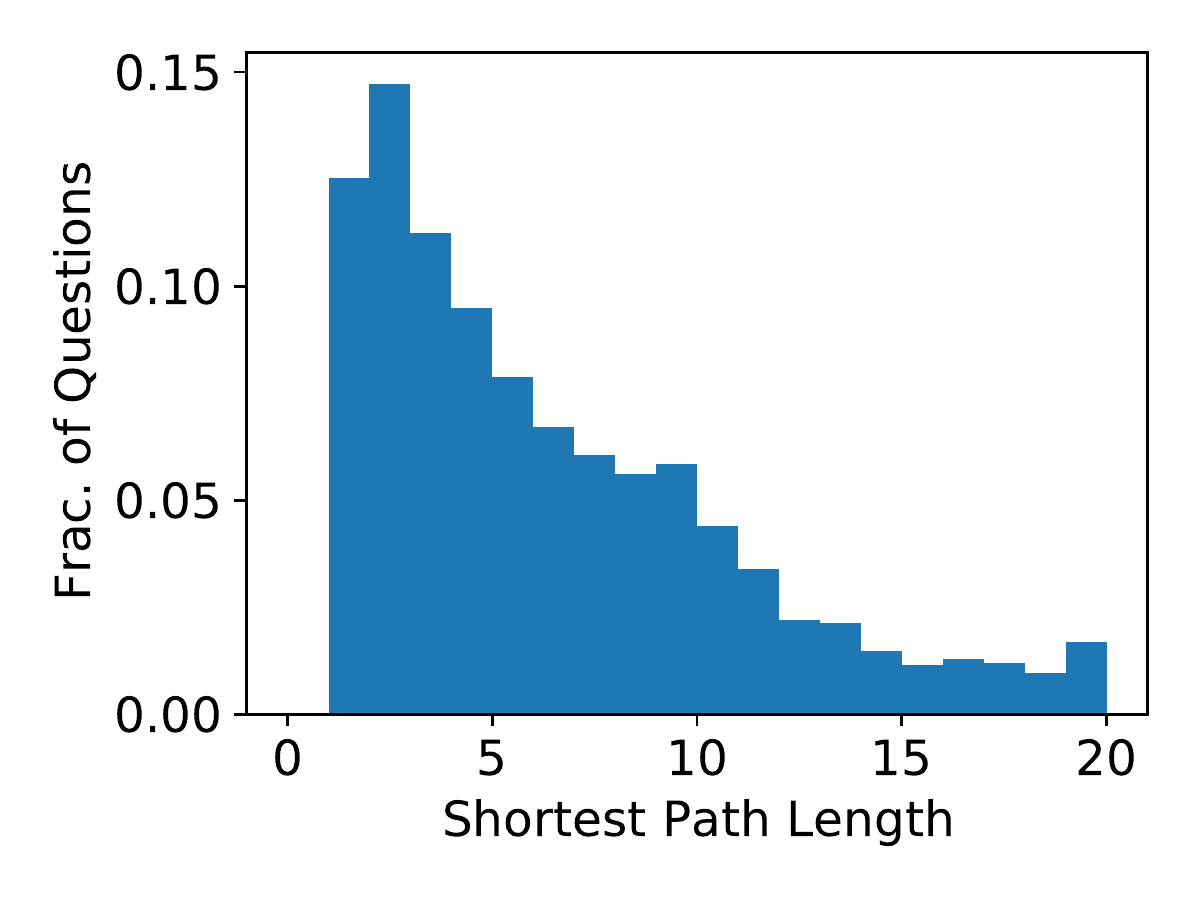}
        \caption{Test/PDF}
        \label{fig:nq-dspl-test-pdf-zoom}
    \end{subfigure}
    \hfill
    \begin{subfigure}{0.33\linewidth}
        \includegraphics[width=\textwidth]{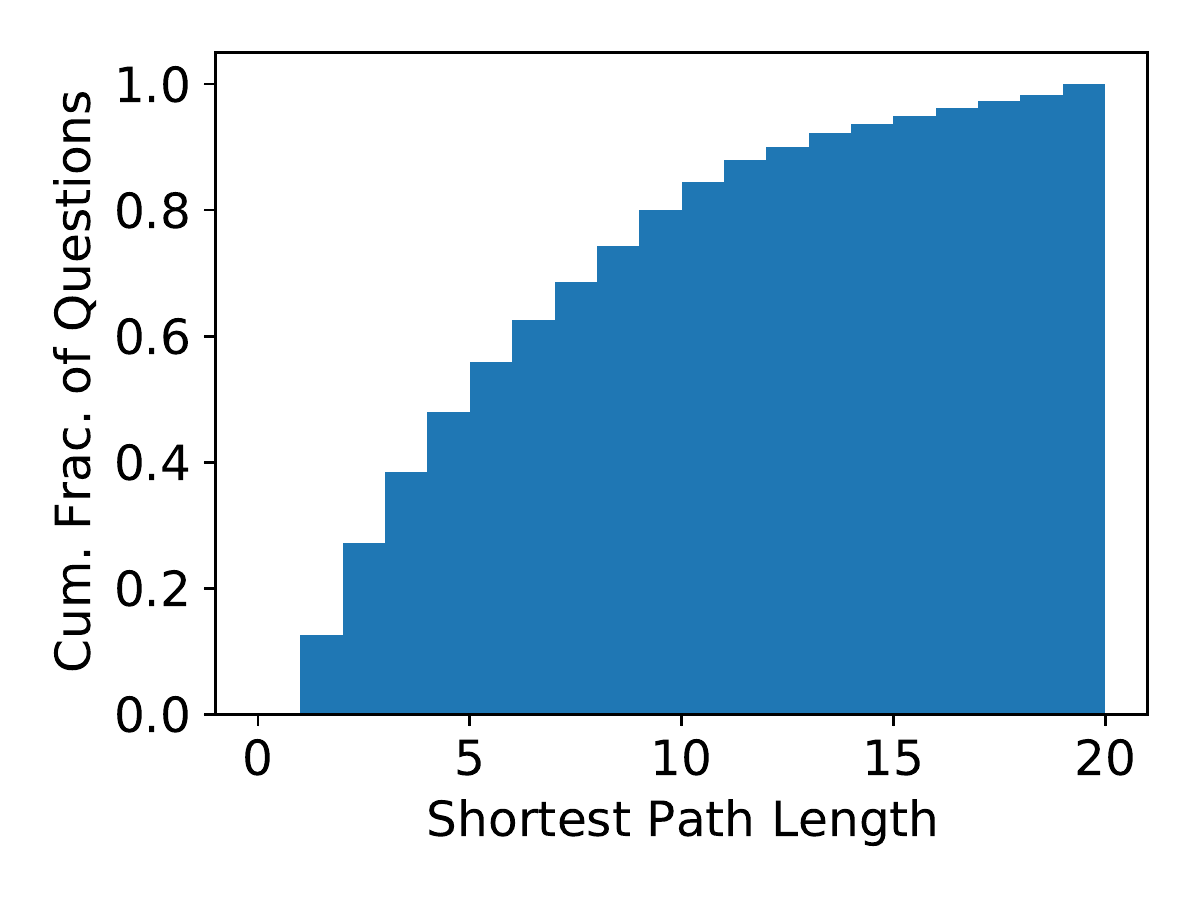}
        \caption{Test/CDF}
        \label{fig:nq-dspl-test-cdf-zoom}
    \end{subfigure}
    \hfill
    \phantom{}\\
    \hfill
    \begin{subfigure}{0.33\linewidth}
        \includegraphics[width=\textwidth]{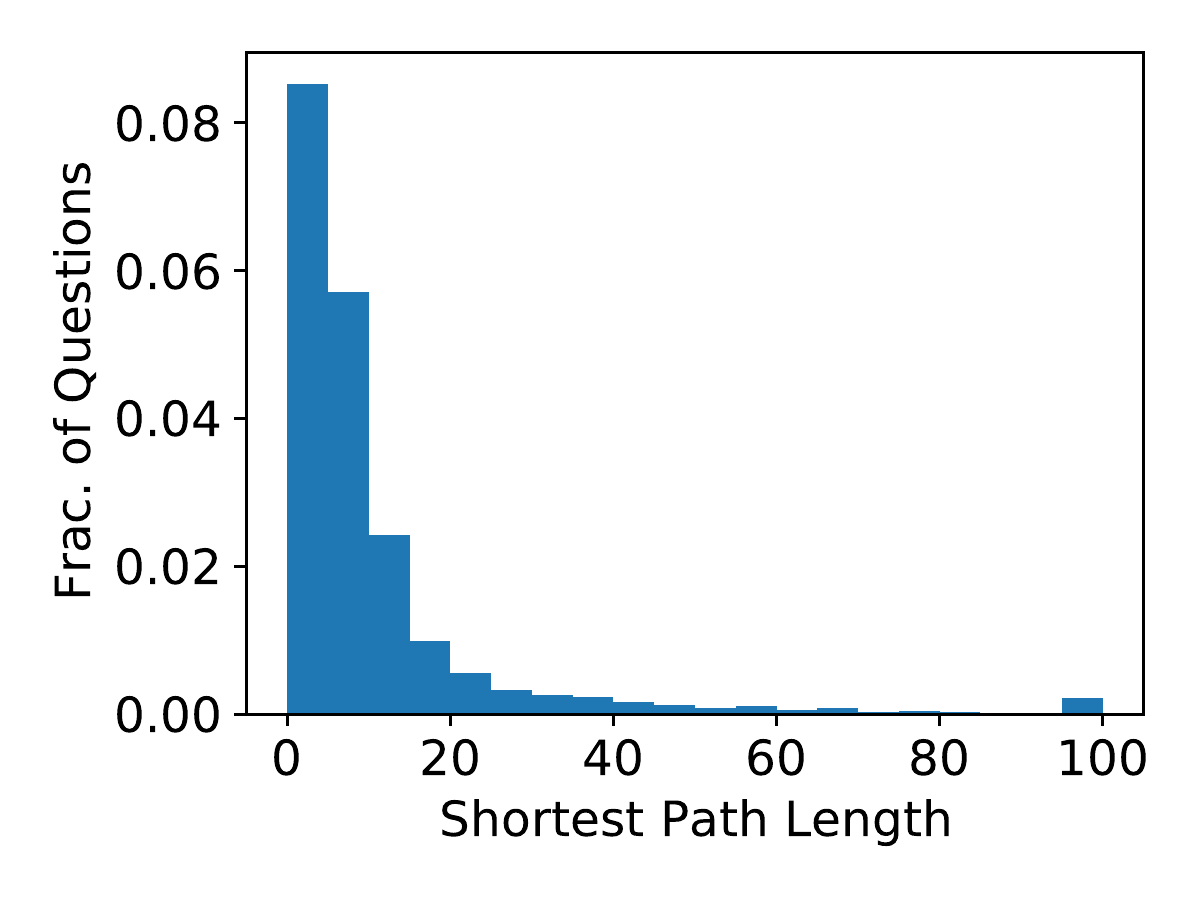}
        \caption{Test/PDF}
        \label{fig:nq-dspl-test-pdf}
    \end{subfigure}
    \hfill
    \begin{subfigure}{0.33\linewidth}
        \includegraphics[width=\textwidth]{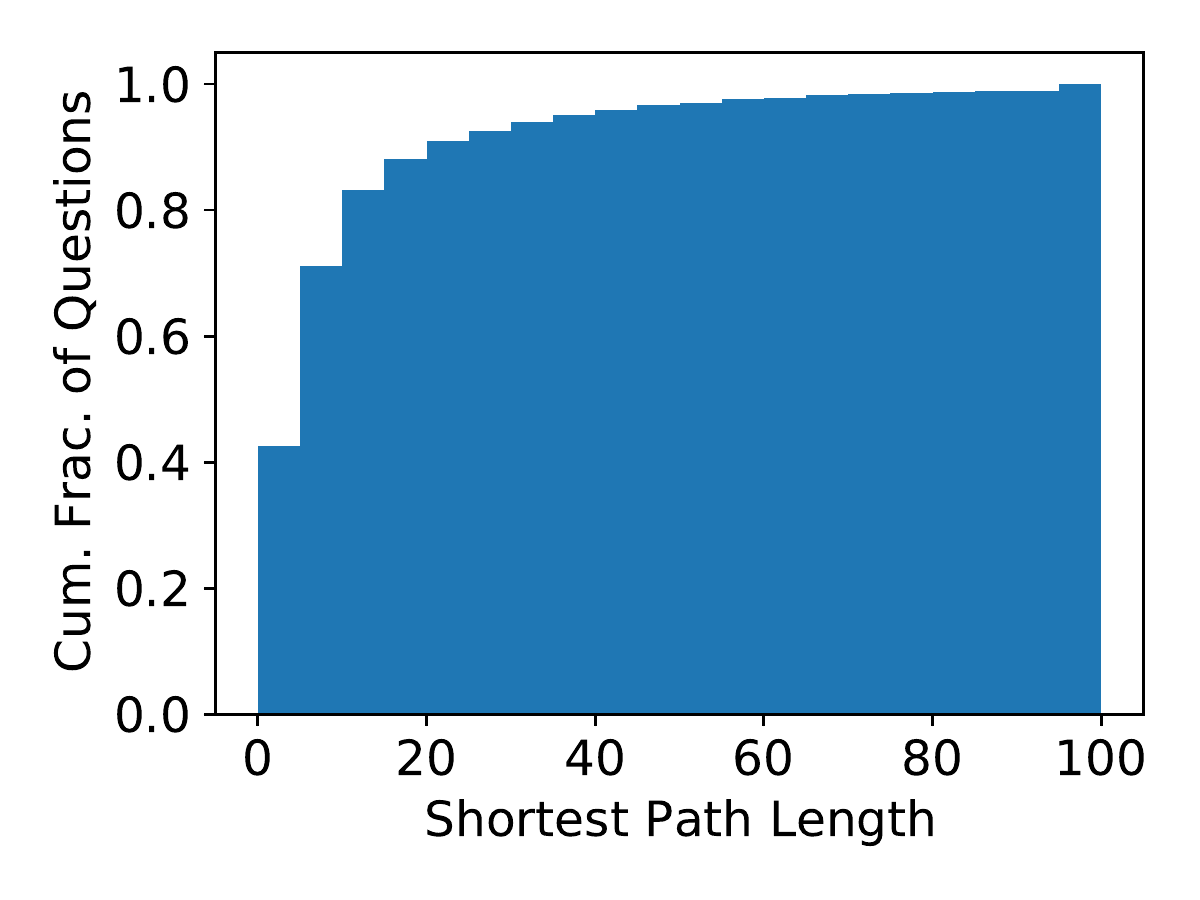}
        \caption{Test/CDF}
        \label{fig:nq-dspl-test-cdf}
    \end{subfigure}
    \hfill
    \vspace{-1mm}
    \caption{Distribution of Short Path Length from starting node computed by BM25 to target node on Natural Questions dataset on 2018 graph.}
    \label{fig:nq-dspl}
\end{figure}

\paragraph{Data statistics}
We compute the shortest-path distance between top-1 retrieval of BM25 and target node in this graph in Figure~\ref{fig:nq-dspl}.
As can be seen that most path lengths are relatively short, which implies that BM25 lands us in the right vicinity and by a small amount of navigation around we will be able to find the right evidence passage.

\paragraph{Training details}
For the NQ benchmark, our model uses learned embeddings for $\phi$ with a single feed-forward layer for the policy network. The model is pretrained on the 
$5$-step sentence search task, and finetuned on trajectories generated from the NQ dataset using the the following loss function: \begin{equation}
    \mathcal{L}(\theta) = \mathcal{L}_{\rm BC}(\theta) + 0.1 \| \phi_\text{target}(\text{question}) - \phi(s_g)   \|_2
\end{equation}
where $\mathcal{L}_{\rm BC}$ is the normal BC loss we use in our other experiments. 

Only the $\phi_\text{target}$ weights are finetuned for this task. We use AdamW with a constant learning rate of $10^{-6}, \beta_1=0.9, \beta_2=0.999$, and $\epsilon=10^{-6}$. The model is finetuned for $100,000$ update steps with a batch size of $512$. No edge dropout is used during finetuning.

\paragraph{Evaluation details}
For each question, we first run BM25 over all nodes in the graph and take the top-5 matches as starting nodes. 
From each starting node, the agent then navigates for 20 steps.
All the 100 nodes visited by the agent is then ranked by a simple cross-attention model.
For the cross-attention model, we follow~\citet{hofstatter2020improving} to train 6-layer BigBird~\citep{zaheer2020bigbird}.
Since we target navigating to the exact evidence passage required to answer the question, we use recall@\{1,2,3,4,5\} for finding the gold evidence passage as our metric.
More commonly the metric marks a retrieved passage to be correct if it contains the answer string, but this causes a lot of false positives, e.g. the answer string appears in a totally irrelevant context.

\newpage

\section{Variational Interpretation of our Approach}
\label{app:varinterp}
We present an alternative %
motivation of our method based on variational inference in a latent variable model of states. Assume we are given some goal node $n_g$. We would like to parameterize a model $p_\theta(n_T| n_g)$ which will generate a trajectory of states $n_0,n_1,\ldots,n_T$ such that $n_T = n_g$. We define our model as a latent variable model
\begin{align}
    p_\theta(n_T | n_g) &= \sum_{n_0,n_1,\ldots, n_{T-1}} p_\theta(n_T, n_{T-1}, \ldots, n_1, n_0 | n_g )\nonumber\\
    &= \sum_{n_0,n_1,\ldots, n_{T-1}} p(n_0) \prod_{t=1}^T p_\theta(n_t | n_{t-1}, n_g )
\end{align}
i.e an autoregressive model which samples an initial state from $n_0 \sim p(n_0)$ and then samples subsequent states $n_t \sim p_\theta(n_t | n_{t-1}, n_g)$. The transition distribution samples the next node $n_t$ from the neighbors of $n_{t-1}$ in our graph. We would like the probability that $n_T = n_g$ to be large, thus we will train our model to maximize $\log p_\theta(n_T=n_g | n_g)$.

For latent-variable models such as this, we can rewrite the marginal likelihood as 
\begin{align}
    \log p_\theta(n_T | n_g) = \mathbf{E}_{p_\theta(n_0, \ldots, n_{T-1}|n_T, n_g)}  \left[p(n_0) \prod_{t=1}^T p_\theta(n_t | n_{t-1}, n_g ) - \log p_\theta(n_0, \ldots, n_{T-1}|n_T, n_g)\right]
\end{align}
and can obtain a lower-bound on this quantity by replacing the intractable posterior $p_\theta(n_0, \ldots, n_{T-1}|n_T, n_g)$ with a variational approximation $q(n_0, \ldots, n_{T-1}|n_T, n_g)$, i.e
\begin{align}
    \mathcal{L}_\theta(n_T, n_g; q) &:= \log p_\theta(n_T | n_g)\nonumber\\
    &\geq \mathbf{E}_{q(n_0, \ldots, n_{T-1}|n_T, n_g)}  \left[p(n_0) \prod_{t=1}^T p_\theta(n_t | n_{t-1}, n_g ) - \log q(n_0, \ldots, n_{T-1}|n_T, n_g) \right].
\end{align}

The above bound becomes tight when $q(n_0, \ldots, n_{T-1}|n_T, n_g) = p_\theta(n_0, \ldots, n_{T-1}|n_T, n_g)$. Thus, we can optimize our model parameters $\theta$ to maximize $\mathcal{L}_\theta(n_T, n_g; q)$. This approach has had a long history of successfully training latent-variable generative models~\citep{kingma2013auto, ho2020denoising}.

In the context of this work, we can view $q(n_0, \ldots, n_{T-1}|n_T, n_g)$ as a distribution over trajectories of nodes which end at our goal node. We can interpret the various trajectory generation methods introduced in Section~\ref{sec:trajdist} as different variational approximations $q$. Ideally, the option which most closely approximates the true posterior $p_\theta(n_0, \ldots, n_{T-1}|n_T, n_g)$ would be the most desirable but in general this distribution is intractable. 

We find that using simple random trajectories provides a good-enough approximation to the posterior to enable us to train a model which reliably finds the goal state. This result is not completely surprising in context of prior work~\citep{dai2020learning} which successfully trains latent-variable models using random trajectories as an inference model.

\section{Efficiency Analysis}
\label{sec:efficiency}
In this section we analyze the asmpytotic runtime of our method and baselines. In Table~\ref{tab:efficiency} we show the asymptotic runtime and whether or not it is scalable for each method. We also show the 5-step accuracy from Table~\ref{tab:simplenavigation} for quick reference.

\begin{table}[ht]
    \centering
    \caption{Asmptotic runtime analysis of RFBC and baselines.}
    \label{tab:efficiency}
    \centering
    \begin{tabular}{@{}lllc@{}}
    \toprule
    Method & Cost (Big-O) & Scalable & Accuracy (5-step 200k graph) \\
    \\
    \midrule
    RFBC (ours) & $O(E_{out} T)$ & Yes & 85.3 \\
    Backwards BC & $O(E_{out} T)$ & Yes & 2.3 \\
    Shortest path BC & $O(E \log V + E_{out} T)$ & No & 86.7 \\
    RL & $O(E_{out} T)$ & Yes & 41.4 \\
    Random& $O(T)$ & Yes & 12.3 \\
    Greedy & $O(E_{out} T)$ & Yes & 19.7 \\
    Random DFS & $O(T)$ & Yes & 10.0 \\
    Greedy DFS & $O(E_{out} T)$ & Yes & 31.1 \\
    \bottomrule
  \end{tabular}
  \\
  \begin{tabular}{ll}
  $V$ & nodes in the graph [large -- $10^7$] \\
  $E$ & total edges in the graph  [very large -- $10^8-10^9$] \\
  $E_{out}$ & average out degree of node (\# of actions)  [small -- $10^1$] \\
  $T$ & length of trajectory, i.e. the maximum steps agents can run. [small -- $10^1$] \\
  \end{tabular}
\end{table}

$T$ is the trajectory length (either the length of the trajectory for BC methods or the maximum allowed trajectory for other methods. For BC, $T$ is guaranteed to be shorter than the max, but this is a constant factor difference.
$E_{out}$ is the average number of outgoing edges (i.e. actions) at a node. (small - $10^1$)
$E$ is the total edges in the graph (very large - $10^8$).
$V$ is the number of nodes in the graph (large - $10^7$).

For most methods, the runtime is $O(E_{out} T)$ because the method does some constant amount of work for each possible actions $E_out$ for $O(T)$ steps. For the random methods it is $O(T)$ because the work is constant per step (choose randomly). Shortest path distance has the worst runtime. The $O(E log V)$ term is the runtime of finding a shortest path (Dijkstra's algorithm with Fibonacci heap). As the graph gets larger, this term because infeasible (we could not even run this on the full Wikipedia graph). DFS might seem like it should be larger than $O(E_{out} T)$ or $O(T)$, but because we restrict the maximum allowable steps, it is only proportional to $T$ and (for greedy) $E_{out}$.

From this Table, we see that all methods (except shortest path) have scalable runtimes. 
The dominant factor is $E_{out}$, which theoretically grows slowly with graph size.
Moreover, in practice, there is a natural upper bound on the number of links one can have in a paragraph of text, and the runtime difference in our experiments going from 200k nodes to 38M nodes is only 1.5x despite the 200-fold increase in graph size.

\end{document}